\documentclass{article}

\usepackage{microtype}
\usepackage{graphicx}
\usepackage{subfigure}
\usepackage{booktabs} 
\usepackage[table]{xcolor}
\usepackage{hyperref}

\usepackage[accepted]{icml2024}

\usepackage{amsmath}
\usepackage{amssymb}
\usepackage{mathtools}
\usepackage{amsthm}

\usepackage[normalem]{ulem}

\usepackage{booktabs}
\usepackage{multirow}
\usepackage{amsfonts}
\usepackage{breqn}
\def\leref#1{Lemma~\ref{#1}}

\DeclareMathOperator*{\argmin}{arg\,min\,}

\newcommand{\methodname}{{\tt{FedNSL}}}

\usepackage[capitalize,noabbrev]{cleveref}

\theoremstyle{plain}
\newtheorem{theorem}{Theorem}[section]

\newtheorem{lemma}[theorem]{Lemma}

\theoremstyle{definition}

\theoremstyle{remark}

\usepackage[textsize=tiny]{todonotes}

\begin{document}

\twocolumn[
\icmltitle{Federated Neuro-Symbolic Learning}



\icmlsetsymbol{equal}{*}

\begin{icmlauthorlist}
\icmlauthor{Pengwei Xing}{ntu}
\icmlauthor{Songtao Lu}{ibm}
\icmlauthor{Han Yu}{ntu}

\end{icmlauthorlist}

\icmlaffiliation{ntu}{College of Computing and Data Science, Nanyang Technological University, Singapore}
\icmlaffiliation{ibm}{IBM Thomas J. Watson Research Center Yorktown Heights, USA}

\icmlcorrespondingauthor{Pengwei Xing}{pengwei001@e.ntu.edu.sg}
\icmlcorrespondingauthor{Songtao Lu}{songtao@ibm.com}
\icmlcorrespondingauthor{Han Yu}{han.yu@ntu.edu.sg}

\icmlkeywords{Machine Learning, ICML}

\vskip 0.3in
]



\printAffiliationsAndNotice{\icmlEqualContribution} 

\begin{abstract}

Neuro-symbolic learning (NSL) models complex symbolic rule patterns into latent variable distributions by neural networks, which reduces rule search space and generates unseen rules to improve downstream task performance. Centralized NSL learning involves directly acquiring data from downstream tasks, which is not feasible for federated learning (FL). To address this limitation, we shift the focus from such a one-to-one interactive neuro-symbolic paradigm to one-to-many \underline{Fed}erated \underline{N}euro-\underline{S}ymbolic \underline{L}earning framework (\methodname{}) with latent variables as the FL communication medium.
Built on the basis of our novel reformulation of the NSL theory, \methodname{} is capable of identifying and addressing rule distribution heterogeneity through a simple and effective Kullback-Leibler (KL) divergence constraint on rule distribution applicable under the FL setting. It further theoretically adjusts variational expectation maximization (V-EM) to reduce the rule search space across domains. This is the first incorporation of distribution-coupled bilevel optimization into FL. Extensive experiments based on both synthetic and real-world data demonstrate significant advantages of \methodname{} compared to five state-of-the-art methods. It outperforms the best baseline by 17\% and 29\% in terms of unbalanced average training accuracy and unseen average testing accuracy, respectively.
\end{abstract}

\section{Introduction}
Neuro-symbolic learning (NSL) \cite{garcez2008neural} stands at the frontier of artificial intelligence, amalgamating symbolic reasoning with the prowess of neural networks. Neuro-symbolic work is generally divided into two categories: one focuses on concept extraction, mapping the neural network's inter-node structure into hierarchical relationships, for instance, by using label hierarchies from label classification \cite{ciravegna2023logic}. This method covertly transforms the network relationships into specific concepts through methods such as binarization or truth table comparisons \cite{ciravegna2020constraint}. The other category predominantly involves knowledge graph (KG) \cite{rebele2016yago, Liu-et-al:2024TNNLS}, concentrating on learning semantic logic rules from natural language among KG. These approaches often employ sequence models like Transformers to extract these logic relationships \cite{qu2020rnnlogic,xu2022ruleformer} for KG-related downstream tasks, such as KG completion \cite{bordes2013translating,wang2014knowledge}, relation extraction \cite{weston2013connecting,riedel2013relation} and entity classification \cite{nickel2011three,nickel2012factorizing}.
These two methods transform the representations learned by neural networks into first-order logic (FOL) systems. Through this transformation, they can use a unified symbolic form to interpret, further infer, and optimize the representations learned by the neural networks.

\begin{table*}[t]
\centering
\resizebox{0.9\textwidth}{!}{%
\begin{tabular}{@{}llllll@{}}
\toprule
PFL &
  \begin{tabular}[c]{@{}l@{}}Representative\\ Work\end{tabular} &
  Server Objective &
  Local Constraint &
  \begin{tabular}[c]{@{}l@{}}Communication\\ Medium\end{tabular} &
  \begin{tabular}[c]{@{}l@{}}Problem\\ Dimension\end{tabular} \\ \midrule
Regularization Based & pFedMe     & Weight Generalization & \begin{tabular}[c]{@{}l@{}}Difference on\\ Weights\end{tabular}    & Weight      & Same      \\
Meta-learning Based  & Per-FedAvg & Gradient of Gradient  & \begin{tabular}[c]{@{}l@{}}Adjustments on\\ New Data\end{tabular}  & Weight      & Same      \\
Bayesian Based       & pFedBayes  & Prior of Weight       & \begin{tabular}[c]{@{}l@{}}KL-Divergence on\\ Weights\end{tabular} & Weight      & Same      \\
Rule-Alignment Based & LR-XFL     & Rule Alignment        & Quality of Rule                                                      & Rule, Weight & Same \\
\midrule
\rowcolor{purple!8} Neuro-Symbolic Based ({\bf this work}) &
  \methodname{} &
  Rule Distribution &
  \begin{tabular}[c]{@{}l@{}}KL-Divergence on\\ Rule\end{tabular} &
  \begin{tabular}[c]{@{}l@{}}Distribution\\ Variance\end{tabular} &
  Different \\ \bottomrule
\end{tabular}%
}
\caption{Our framework stands apart from other approaches in that its problem dimensions vary, making it suitable for a broader range of federated scenarios. Unlike conventional methods, the decision variables in the optimization objectives for the server and the client are not the same. This allows for the utilization of entirely different task types to aid the local models, as opposed to merely focusing on aggregating weights of identical dimensions.}
\label{tb:pfl_comp}
\vspace{-4mm}
\end{table*} 

\begin{figure*}[!t]
\centering
\includegraphics[width=0.90\textwidth]{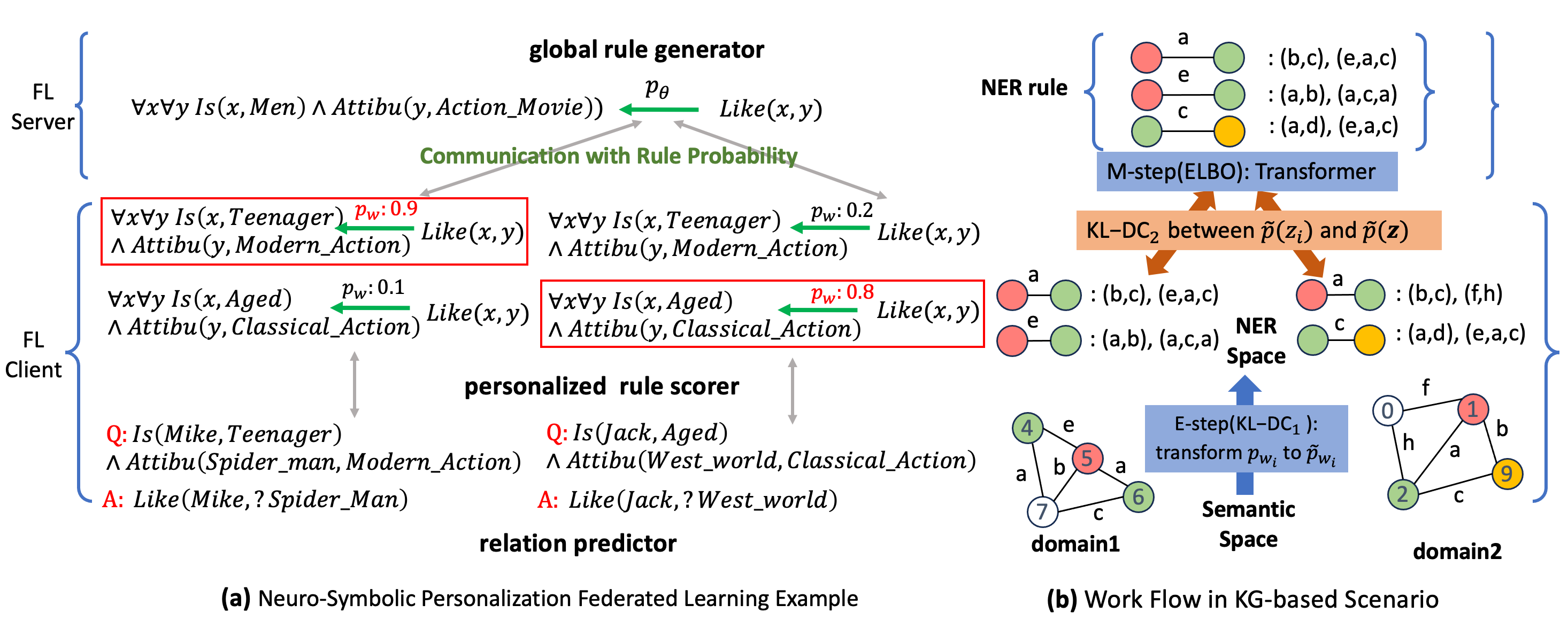}
\caption{A neuro-symbolic PFL example (a) and a corresponding KG-based rule learning scenario (b). The example (a) illustrates how a global rule generator and multiple personalized rule scorers cooperate to tackle rule personalization without exposing local data by transmitting rule distribution probabilities. Meanwhile, the KG-based workflow (b) demonstrates the V-EM mechanism, employing maximization of $\mathrm{ELMO}$ and minimization of $\mathrm{KL}$-divergence constraint-1 (KL-DC1) for inductive rule reference (blue part in (b)). Additionally, it incorporates a $\mathrm{KL}$-divergence constraint-2 (KL-DC2) to diminish rule heterogeneity (orange part in (b)).}
\label{fig:two_flow}
\vspace{-4mm}
\end{figure*}

Personalized requirements for NSL can also be reflected through the FOL symbolic form. Considering a scenario of personalized movie recommendations based on user preferences. If there are two groups of people, the aged group of men and the teenage group of men, there should be a generality and a specificity in the degree of preference for the type of movie. The more general logic rule that the neuro-symbolic system can learn is $\forall X \forall Y ((is(X, Men) \wedge attribute(Y, Action\_Movie))\rightarrow like(X, Y))$. The more personalized logic rule will be $\forall X \forall Y ((is(X, Teenager) \wedge attribute(Y, Modern\_Action)) \rightarrow like(X, Y))$ for the teenager group of men and $\forall X \forall Y ((is(X, Aged) \wedge attribute(Y, Classical\_Action)) \rightarrow like(X, Y))$ for the aged group of men, as shown in Figure \ref{fig:two_flow} (a). Due to the heterogeneity of the rule from data heterogeneity, it is crucial 
 to seek a trade-off between personalization and generalization across multiple domains. Moreover, privacy has emerged as a critical concern with the rise of federated learning (FL) \cite{tan2022towards,Goebel-et-al:2023}. In the above scenario, the system utilizes first-order logic to capture individual user preferences for movies. These preferences may contain sensitive information about users' tastes, interests, or potentially even personal characteristics inferred from their movie choices. Therefore, the personalized data is not shareable or available during the learning process on the server. This motivates us to study the following question.

\textit{Can we realize neuro-symbolic learning over the heterogeneous federated setting?}

In this paper, we propose a new distributed framework, named \methodname{}, for federated NSL based on interplaying prior rule distribution learning on the FL server with many downstream-task-related posterior rule distribution learnings on the FL client. Leveraging distribution-coupled bilevel optimization (BO) \cite{lu2023bilevel} (rather than the traditional weight-coupled BO \cite{dickens2024convex}), 
we re-formalize neuro-symbolic learning within an FL context, enabling the identification of heterogeneity as originating from latent variable distributions of rules between FL server and clients. We further solve this distribution-coupled problem using the V-EM that has been tailored for the federated NSL setting.

 The major contributions of this work are as follows:
\begin{itemize}
    \item  To the best of our knowledge, this is the first framework that addresses the heterogeneity issue of rule induction under FL settings, and the first time that a distribution-coupled BO problem has been addressed under FL settings.
    \item  We theoretically propose a factorizable federated V-EM algorithm to solve coupled distribution and significantly enhance the search efficiency of the huge rule space in cross-domain scenarios.
    \item Under stringent cross-visible rule distribution experiment settings, \methodname{} outperforms the best baseline by 17\% and 29\% in terms of unbalanced average training accuracy and unseen average test accuracy, respectively.
\end{itemize}

\vspace{-2mm}
\section{Related Works}

\subsection{Neuro-Symbolic Learning}
NSL represents a cutting-edge fusion in artificial intelligence, blending the learning capabilities of neural networks (NNs) with the structured reasoning of symbolic logic. Integration appears in diverse forms, such as extracting logic concepts by binarizing and pruning neural networks \cite{ciravegna2023logic}, embedding logical structure within neural network frameworks \cite{yang2017differentiable}, and leveraging logical rules as constraints for regularizing neural networks \cite{ciravegna2020constraint,9415044}. Additionally, logic inference with differentiable networks \cite{minervini2020learning} or probabilistic models \cite{qu2020rnnlogic} allows for inductive rule learning to produce many generative logical relationships for complex, real-world relationships.

A particularly active and significant branch within NSL is the use of KGs for semantic logic reasoning \cite{zhang2021neural}. This approach stands out due to its scalability in handling large datasets and its ability to infer new logic knowledge. KG-based NSL falls under the umbrella of inductive logic learning, leveraging the power of KGs to learn semantic logical relationships as latent variables with sequence models like Transformer \cite{nafar2023teaching,ru2021learning,qu2020rnnlogic}. This method enables the generation of unseen logical relations, significantly reducing the search space within large-scale knowledge graphs. However, dealing with the rule heterogeneity of latent variables brought about by data heterogeneity remains a challenge.

\vspace{-2mm}
\subsection{Personalized Federated Learning}

Personalized federated learning (PFL) \cite{tan2021towards} emphasizes aggregating models with privacy protection and focuses on solving the issue of data heterogeneity in various scenarios of federated learning using various methods. Regularization-based PFL seeks this balance through a formulation that minimizes a difference function combining local loss and a regularization term, linking local models to a global standard, as seen in methods like FedU \cite{dinh2021fedu}, pFedMe \cite{t2020personalized}, and FedAMP \cite{huang2021personalized}. Meta-learning based PFL, represented by pioneering works like Per-FedAvg \cite{fallah2020personalized}, involves a two-step process where the meta model is realized by weight aggregation, followed by each client fine-tuning on new batch data with the gradient of gradient (i.e., second-order information of the loss function). Bayesian-based PFL, illustrated by pFedGP \cite{achituve2021personalized} and pFedBayes \cite{zhang2022personalized}, adopts a probabilistic approach by considering the global weight as a prior distribution of all local weights and using a KL-divergence on weight distribution as a constraint. 

For federated NSL, the objective is to address the rule heterogeneity. Although rule heterogeneity originates from data heterogeneity, the approach in NSL needs to consider the varying degrees of rule uncertainty across different clients and avoid extensive rule transmission due to the vast rule search space. Although LR-XFL \cite{zhang2023lr} attempts to mitigate the rule heterogeneity by solving the rule conflict in rule alignment and assigning different proportions for global weight aggregation for different clients based on the quality of rules. However, it cannot avoid exhaustive rule searching in the process of transferring and aligning rules, leading to inefficient communication. Furthermore, any change in local data might necessitate the re-extraction and realignment of rules, thus failing to effectively handle the uncertainty of rules.

\vspace{-2mm}
\section{The Proposed \methodname{} Approach}

In our study, we investigate a distribution-coupled bilevel optimization framework, specially developed for federated NSL. This framework is crafted to address multiple challenges: it reduces rule heterogeneity, boosts communication efficiency, and decreases rule uncertainty. The enhancement in communication efficiency is achieved by narrowing the rule search space, while the reduction in rule uncertainty is accomplished by learning a rule distribution on the server. This distribution serves as a means of personalization in FL for local tasks. Further, leveraging the power of generative probabilistic models, the proposed framework is able to help with providing a diverse set of samples following rule distributions, intending to avoid overfitting issues during the training process. (Table~\ref{tb:pfl_comp} is a summary of the differences between ours and other PFL ways.)

\vspace{-2mm}
\subsection{Overview of \methodname{}}

Figure~\ref{fig:two_flow} (b) illustrates our workflow, which utilizes a V-EM algorithm that integrates prior rule distribution learning on the FL server with $n$ downstream-task-related posterior rule distributions formed on the FL clients. On the server level, a global transformer-based sequence model is employed for the M-step to learn a rule generator informed by the prior rule distribution. This rule generator produces multiple candidate rule bodies $r_1 \wedge ... \wedge r_l$ corresponding to rule head $r_{\mathrm{head}}$ for each rule category in rule latent variable $z$ space. In a KG context, a rule category might be ``Person-Place'', where any rule bodies with the head $r_1$'s NER as ``Person'' and the tail $r_l$'s NER as ``Place'' are classified under this category. Each client, during the E-step, receives these candidate rule bodies from the server and utilizes a rule scorer $w_i$ to evaluate and select the most suitable rule body from all candidates received. The goal is to enhance the referential representation of $r_{\mathrm{head}}$, thereby improving the prediction accuracy of the relation $a_i$ (equivalent to $r_{\mathrm{head}}$) for $q_i$ in the specific head-relation-tail triplet $\langle h,?,t \rangle$ within knowledge graph $\mathcal{G}_i$. The overarching objective is to refine the rule generator to subsequently enhance the relation predictor for the downstream task.


It is noteworthy that in the details of algorithm~\ref{ag:algorithm}, local data are not exposed by transmitting distributional probabilities in a framework where the prior rule distribution of the server and posterior rule distribution of local are coupled to each other. From a privacy-preservation perspective, this is equivalent to transmitting probabilistic model weights or parameters, aligning with other baselines in Table \ref{tb:pfl_comp}. To be specific, the generation of candidate rule bodies can be achieved by directly sampling with the prior rule distribution probabilities distributed from the server without avoiding rule transmitting. Similarly, the posterior distribution probabilities can also be transmitted to the server, where the server samples new data from these probabilities to incorporate into the transformer training samples.

\vspace{-2mm}
\subsection{Distribution-Coupled Bilevel Optimization Objective for \methodname{}}

To meet the requirement for coping with coupled distribution, we consider a new FL paradigm. In this setting, we can communicate the rule distribution. The new federated neuro-symbolic objective can be formulated in the following bilevel programming form:
\vspace{-1mm}
\begin{subequations} \label{eq:one_many_FL}
\begin{align} \label{eq:one_many_FL.a}  
\min_\theta \; & \mathop{\mathbb{E}}_{\bar{z}\sim p_{\theta}(\cup_{i=1}^{n} w_i^{\star}(\theta))} \mathcal{T}(\theta,\{\cup_{i=1}^{n}{w_i}^{\star}(\theta)\};\bar{z})
\\  \label{eq:one_many_FL.b} 
\textrm{s.t.}\; &
w_i^{\star}(\theta)\in \argmin_{w_i}  
\mathop{\mathbb{E}}_{z_i\sim p_{w_i}(\theta)}\mathcal{L}(w_i,\theta;z_i,\bar{z}) \forall i, 
\vspace{-2mm}
\end{align}
\end{subequations}
where
\vspace{-2mm}
\begin{equation} \label{eq:one_many_FL_lower}
\mathcal{L}(w_i,\theta;z_i,\bar{z})=\!\ell(w_i,\theta;z_i) \!+\!\lambda D_{\mathrm{KL}}(p_{w_i}(z_i) ||p_{\theta}(\bar{z})).
\vspace{-2mm}
\end{equation}
The $\mathcal{T}(\cdot)$ and $\mathcal{L}(\cdot)$ respectively denote the upper-level (UL) and lower-level (LL) loss functions, and $i$ represents the index of each client. Additionally, $p_{\theta}$ refers to the global prior rule distribution parametrized by weight $\theta$, while $p_{w_i}$ denotes the posterior rule distribution on the client $i$ parametrized by the weight $w_i$.

In the new context of FL, the tasks of the FL server and the local clients correspond to the  UL  and  LL problems in bilevel optimization, respectively. In this coupled decision-dependent FL scenario, the distribution learned by the rule generator with Eq.~\eqref{eq:one_many_FL.a} on the FL server is dependent on the decision variable $w_i$ on each client in with Eq.~\eqref{eq:one_many_FL.b} (denoted by $\bar{z}\sim p_{\theta}(\cup_{i=1}^{n} w_i^{\star}(\theta))$). This dependency arises because the downstream tasks need to update $w_i$ using label $a_i$ and $\mathcal{G}_i$ of local data to form a posterior probability, which, once uploaded to the server, impacts the server's prior distribution $p_{\theta}$ and global latent rule $\bar{z}$. Simultaneously, inference and prediction data for downstream tasks are dependent on the distributed rules from the server's rule generator with the decision variable $\theta$ (denoted by $z_i\sim p_{w_i}(\theta)$).

\vspace{-2mm}
\subsection{Rule Distribution Heterogeneity}
It is worth noting that the optimization of the server's objective function is based on the expectation $\mathop{\mathbb{E}}_{\bar{z}\sim p_{\theta}(\cup_{i=1}^{n} w_i^{\star}(\theta))}\mathcal{T}(\cdot)$ w.r.t. the global $\bar{z}$ distribution in Eq.~\eqref{eq:one_many_FL.a}, while the optimization of the client's objective function is based on the expectation $\mathop{\mathbb{E}}_{z_i\sim p_{w_i}(\theta)}\mathcal{L}(\cdot)$ w.r.t. $z_i$ of each client $i$ in Eq.~\eqref{eq:one_many_FL.b}. This discrepancy between them is the rule heterogeneity we research in this paper. Only when all $z_{i}\forall i$ are independent and identically distributed (i.i.d.), $\bar{z}\sim p_{\theta}(\cup_{i=1}^{n} w_i^{\star}(\theta))$ will be i.i.d. with $z_i\sim p_{w_i}(\theta)$ for downstream tasks.  To address this, a KL-divergence-based penalty term $D_{\mathrm{KL}}(p_{w_i}(z_i) ||p_{\theta}(\bar{z}))$ must be added to the objective function of each client, which constrains the divergence between the posterior distribution $p_{w_i}$ of personalized $z_i$ and the prior distribution $p_{\theta}$ of the globally shared $\bar{z}$.

\vspace{-2mm}

\vspace{-1mm}
\subsection{Solving Objective with Variational Expectation Maximization to Reduce Search Space}

In this section, we will provide specific expressions for the objective function of the server and each client in Eq.~\eqref{eq:one_many_FL.a}, Eq.~\eqref{eq:one_many_FL.b} and $D_{\mathrm{KL}}(p_{w_i}(z_i) ||p_{\theta}(\bar{z}))$ in Eq~\eqref{eq:one_many_FL_lower} in a concrete KG-based neuro-symbolic scenario considering reduce rule search space. In this scenario, the search space for latent variable logic rules is extensive, indicating that a vast number of KG paths could correspond to $r_{\mathrm{head}}$ in the form of $r_1 \wedge ... \wedge r_l$, provided that the NER of $r_1$ and the NER of $r_l$ belong to a specific pair combination. Directly utilizing a traditional EM algorithm becomes infeasible in this scenario. While in a traditional EM approach, the posterior can be directly computed at the E-step, it's not feasible when dealing with an extensive logic rule space. In this paper, we theoretically adapt the V-EM approach by employing an approximate variational prior rule distribution $\tilde{p}(\bar{z})$ and an approximate variational posterior rule distribution $\tilde{p}(z_i)$. These approximations, which involve a reduced search space, are used in place of the true prior and posterior rule distributions. We use $p_{\theta,w_{1:n}}(\bar{z})$ and $p_{w_{i},\theta}(z_{i}|q_{i},a_{i},\mathcal{G}_{i})$ to denote the true prior and posterior rule distribution in KG scenario for server and client $i$ respectively. Let first discuss the case that $z_i, \forall i$ are i.i.d. with $\bar{z}$, and the non-i.i.d. case (heterogeneous case) will be discussed in Section \ref{sec_cross}.
\begin{lemma}\label{le.log_to_plus_iid}
Given that $z_i, \forall i$ are i.i.d. with $\bar{z}$, the overall log-likelihood function  $\log \left( p_{w_{1:n},\theta}(a_{1:n}|q_{1:n},\mathcal{G}_{1:n}) \right)$ can be rewritten as 
\vspace{-2mm}
\begin{align} \label{eq:vi_iid} \notag
&\quad\;\log \left( p_{w_{1:n},\theta}(a_{1:n}|q_{1:n},\mathcal{G}_{1:n}) \right)
\\\notag
&=\mathcal{L}_{\mathrm{ELBO}}(\tilde{p}(\bar{z}),p_{\theta,w_{1:n}}(\bar{z})) \\
&\quad\quad\quad\quad\quad+\sum_{i=1}^{n}D_{\mathrm{KL}}(\tilde{p}(z_{i}) || p_{w_{i},\theta}(z_{i}|q_{i},a_{i},\mathcal{G}_{i})),
\vspace{-2mm}
\end{align}
where $\mathcal{L}_{\mathrm{ELBO}}(\tilde{p}(\bar{z}),p_{\theta,w_{1:n}}(\bar{z}))$ is the evidence lower bound (ELBO) of the log-likelihood function, and $D_{\mathrm{KL}}(\tilde{p}(z_i)||p_{w_{i},\theta}(z_i|q_{i},a_{i},\mathcal{G}_{i}))$ is the KL-divergence between approximate posterior distribution and true posterior distribution on each client $i$. In addition, maximizing the overall log-likelihood function is achieved by maximizing the shared $\mathcal{L}_{\mathrm{ELBO}}(\tilde{p}(\bar{z}),p_{\theta,w_{1:n}}(\bar{z}))$ on the FL server and minimizing $D_{\mathrm{KL}}(\tilde{p}(z_i)||p_{w_{i},\theta}(z_i|q_{i},a_{i},\mathcal{G}_{i}))$ on each FL client $i$.
\end{lemma}
Due to the page limit, the detailed proofs of all the lemmas in the main text are delegated in the appendix.


\vspace{-1mm}
\subsubsection{M-Step of V-EM on Server}

At the upper server level, given \leref{le.log_to_plus_iid}, we can maximize the lower bound $\mathcal{L}_{\mathrm{ELBO}}(\tilde{p}(\bar{z}), p_{\theta,w_{1:n}}(\bar{z}))$ with the M-step of the V-EM algorithm.
\begin{lemma}\label{le.mstep}
Given that $z_i, \forall i$ are i.i.d. with $\bar{z}$, maximizing $\mathcal{L}_{\mathrm{ELBO}}(\tilde{p}(\bar{z}),p_{\theta,w_{1:n}}(\bar{z}))$ can be converted into maximizing $\mathbb{E}_{\tilde{p}(\bar{z})}\log\;p_\theta(\bar{z})$ on the FL server, namely,
\begin{align} \label{eq:max_log_mstep}
\max_{\theta}\mathbb{E}_{\tilde{p}(\bar{z})}\log\;p_\theta(\bar{z}).
\end{align}
\end{lemma}
\vspace{-4mm}

Given \leref{le.mstep}, the Eq.~\eqref{eq:one_many_FL.a} in the distribution-coupled bilevel FL paradigm on the server can be converted into maximizing $\mathbb{E}_{\tilde{p}(\bar{z})}\log\;p_\theta(\bar{z})$.
Furthermore, we observe an expectation operation concerning $\tilde{p}(\bar{z})$. In the KG context, by minimizing the loss function $\mathcal{L}_{T_\theta}$, sequence models like a transformer $T_\theta(r_1\wedge ...\wedge r_l|r_{\mathrm{head}})$ can generate multiple candidate rule bodies $r_1\wedge ...\wedge r_l$ for each rule head $r_{\mathrm{head}}$ under a specific distribution. This distribution is specified by the head-NER and tail-NER combination category and rule head $r_{\mathrm{head}}$. Different rule bodies $r_1\wedge ...\wedge r_l$ can be sampled from this distribution. As shown in Figure~\ref{fig:exp_overall} (b), there are three NER categories, each distinguished by different color combinations of nodes. Each category specifies a kind of distribution to which multiple path samples belong. Therefore, we assume the formation of a multinomial distribution w.r.t. $\bar{z}$ can denote a specific NER category distribution, denoted as follows:
\begin{align} \label{eq:mu_theta}
    \tilde{p}_{\theta}(\bar{z}) \sim \mathrm{Mu}_{\theta}(\bar{z}|N,T_{\theta}(r_1\wedge ...\wedge r_l|r_{\mathrm{head}})),
\end{align}
where $\mathrm{Mu}_{\theta}$ denotes the multinomial distribution,  $\tilde{p}_{\theta}(\bar{z})$ stands for the parameterization of prior approximate rule distribution for $\tilde{p}(\bar{z})$ in $\mathcal{L}_{\mathrm{ELBO}}(\tilde{p}(\bar{z}),p_{\theta,w_{1:n}}(\bar{z}))$, and $N$ is the size of the $\bar{z}$. Therefore, the above rule generation process is equivalent to a rule generator performing $N$ samplings to form $J$ unique rule body samples under a multinomial distribution related to the rule head. Consequently, Eq.\eqref{eq:max_log_mstep} can further be written as
\begin{align} \label{eq:max_p_theta}
\max_{\theta}\log \tilde{p}_\theta(\bar{z}).
\end{align}



\vspace{-4mm}
\subsubsection{E-Step of V-EM on Client}
\vspace{-1mm}

At the lower client level, given \leref{le.log_to_plus_iid}, we can minimize $D_{\mathrm{KL}}(\tilde{p}(z_i)||p_{w_{i},\theta}(z_i|q_{i},a_{i},\mathcal{G}_{i}))$ for each client $i$.
\begin{lemma}\label{le.estep}
Given that $z_i, \forall i$ are i.i.d. with $\bar{z}$, minimizing the $D_{\mathrm{KL}}(\tilde{p}(\tilde{z})||p_{w_{i},\theta}(\tilde{z}|q_{i},a_{i},\mathcal{G}_{i}))$ can be converted into maximizing $\mathbb{E}_{\tilde{p}(z_i)}\left[\log p_{w_i}(a_i|z_i,q_i,\mathcal{G}_i)\right]$ in each client $i$, i.e.,
\begin{align}\label{eq:iid_low_obj} 
\max_{w_i} \mathbb{E}_{\tilde{p}(z_i)}\left[\log p_{w_i}(a_i|z_i,q_i,\mathcal{G}_i)\right].
\end{align}
\end{lemma}
The key here is how to solve the expectation on the variational distribution of $\tilde{p}(z_i)$ (i.e, the logic rule space for $p_{w_i}(a_i|z_i,q_i,\mathcal{G}_i)$ in Eq.~\eqref{eq:iid_low_obj}). In the KG scenario, $q_i$ in knowledge graph $\mathcal{G}_i$ is the $\langle h,?,t \rangle$, and the $a_i$ is denoted by $r_{\mathrm{head}}$. In the logic space, the rule generator on the server generates $J$ unique rule bodies $r_1\wedge ...\wedge r_l$ corresponding to $r_{\mathrm{head}}$ for each query $\langle h,?,t \rangle$. At the lower level, our goal is to select the best rule body for a given $r_{\mathrm{head}}$. Then, we shall go through each $r$ from $1$ to $l$ along the path of this rule body $r_1\wedge ...\wedge r_l$. Subsequently, we can use their corresponding fuzzy values to obtain a fuzzy value for $r_{\mathrm{head}}$ to improve relation prediction. Therefore, the likelihood of distribution of answer $a_i$ ($r_{\mathrm{head}}$) is related to all candidate rule bodies. Inspired by \cite{ru2021learning}, we define the likelihood of distribution of $a_i$ in the logic space with the fuzzy values of all candidate rule bodies. According to \leref{le.estep}, minimizing loss function $\ell$ in Eq.~\eqref{eq:one_many_FL_lower} can be converted to maximizing this likelihood as follows:
\vspace{-2mm}
\begin{align}\label{eq:log_pw_KG}\notag
    &\quad\;\mathbb{E}_{\tilde{p}(z_i)} \log \;p_{w_i}(a_i|z_i,q_i,\mathcal{G}_i)
    \\ \notag
    &= \log \sigma\bigg(y(r_{\mathrm{head}})\cdot\!\!\!\!\sum_{\substack{j=1 \\ w_{ij} \in w_i \\z_{ij} \in z_i}}^{J} w_{ij} \cdot \max_{\mathcal{G}_i}\!\!\!\!\!{\prod_{\substack{k=1 \\ r_1\wedge ...\wedge r_l \in z_{ij}}}^{l}\!\!\!\!\!\!\!x(r_{kj})} \bigg),\!\!
    \\
&\approx \frac{1}{2}y(r_{\mathrm{head}})\cdot\sum_{\substack{j=1 \\ w_{ij} \in w_i \\z_{ij} \in z_i}}^{J} w_{ij} \cdot \!\!\!\max_{\mathcal{G}_i}\!\!\!{\prod_{\substack{k=1 \\ r_1\wedge ...\wedge r_l \in z_{ij}}}^{l}\!\!\!x(r_{kj})},
    \vspace{-6mm}
\end{align}
where $\sigma$ denotes the sigmoid function, and $y(r_{\mathrm{head}})$ denotes the relation label of the rule head. We use $j$ to denote the index of all $J$ unique rule bodies in client $i$, and then $w_{ij} \in w_i$ is the learnable weight for the  $j$-th candidate rule body in client $i$. Similarly, $z_{ij} \in z_i$ is the rule distribution variable to denote the $j$-th candidate rule in client $i$. Therefore, $x(r_{kj})$ is the fuzzy value of relation $r_{kj}$ along the path of the rule body $r_1\wedge ...\wedge r_l$ from candidate rule $z_{ij}$. The fuzzy value of relation can be represented by the pre-trained embedding value within the range of $[0,1]$,  $\prod_{k \in l}x(r_{kj})$ denotes the combination of these fuzzy values by multiplying elements along the given path, and $\max_{\mathcal{G}_i}$ denotes the shortest path across $\{\mathcal{G}_i\}$. In the Eq.~\eqref{eq:log_pw_KG}, the second-order Taylor expansion also is applied on $\log p_{w_i}(a_i|z_i,q_i,\mathcal{G}_i)$ to approximate $\log\sigma(\cdot)\approx \frac{1}{2}(\cdot)$ with dropping the constant term $-\log(2)$.

After we have parameterization of $\mathbb{E}_{\tilde{p}(z_i)}\log p_{w_i}(a_i|z_i,q_i,\mathcal{G}_i)$, we define a score function $\mathcal{H}_{w_i}(r_1\wedge ...\wedge r_l|r_{\mathrm{head}})$ for $J$ unique candidate rule bodies as follows:
\begin{align} \label{eq:score_f} \notag
    &\mathcal{H}_{w_i}(r_1\wedge ...\wedge r_l|r_{\mathrm{head}})
    \\ \notag
    &=\frac{1}{J}\cdot \frac{1}{2}y(r_{\mathrm{head}})\cdot \sum_{\substack{j=1 \\ w_{ij} \in w_i \\z_{ij} \in z_i}}^{J} w_{ij} \cdot \max_{\mathcal{G}_i}\!\!\!{\prod_{\substack{k=1 \\ r_1\wedge ...\wedge r_l \in z_{ij}}}^{l}\!\!\!x(r_{kj})}
    \\
    &+ \log(T_{\theta}(r_1\wedge ...\wedge r_l|r_{\mathrm{head}})),
\end{align}
where $T_{\theta}(r_1\wedge ...\wedge r_l|r_{\mathrm{head}})$ is the prior distribution probability from the server which is a constant and $\frac{1}{J}$ is a normalization term.
\begin{lemma}\label{le.tilde_p}
Suppose that $z_i, \forall i$ are i.i.d. with $\bar{z}$ and the  score function is defined in Eq.~\eqref{eq:score_f}, the approximated posterior $\tilde{p}_{w_i}(z_i)$ for each client $i$ is given as follows:
\begin{align} \notag \label{eq:final_tilde_p}
&\tilde{p}_{w_i}(z_i) \propto 
\\
&  \mathrm{Mu}_{w_i}\left(z_i|N,\exp\prod_{j=1}^{J}\left( \mathcal{H}_{w_i}(r_1\wedge ...\wedge r_l|r_{\mathrm{head}}) \right)\right),
\end{align}
where $\tilde{p}_{w_i}(z_i)$ stands for the parameterization of posterior approximate rule distribution for $\tilde{p}(z_i)$ in $D_{\mathrm{KL}}(\tilde{p}(z_{i}) || p_{w_{i},\theta}(z_{i}|q_{i},a_{i},\mathcal{G}_{i}))$
\end{lemma}

With \leref{le.tilde_p}, we can obtain the new posterior $z_i\sim \tilde{p}_{w_i}$ in Eq.~\eqref{eq:one_many_FL.b} which is sent back to the FL server. After that, the FL server draws new posterior samples for the next round of V-EM.

\vspace{-2mm}
\subsubsection{Cross Domain V-EM with Constraint for rule distribution heterogeneity} \label{sec_cross}

\begin{algorithm}[!b]
\caption{\methodname{}}
\label{ag:algorithm}
\begin{algorithmic}[1]
\STATE Initialize
\FOR{ round $k=0,1,2...,K$ }
\STATE //On each FL client:
\FOR{ node $i=0,1,2...,n$ }
\STATE Receive shared prior probabilities $T_{\theta}(r_{\mathrm{head}})$ from the FL server to build a rule distribution and form $J$ unique rule bodies $r_1\wedge ...\wedge r_l$ with it.
\STATE Score these candidate rule bodies with Eq.~\eqref{eq:score_f}.
\STATE Update $w_i$ to solve Eq.~\eqref{eq:one_many_FL.b} by minimizing Eq.~\eqref{eq:one_many_FL_lower} with maximizing Eq.~\eqref{eq:log_pw_KG} using Eq.~\eqref{eq:min_w_add_kl_spec} as regularization term.
\STATE Update new rule's posterior by Eq.~\eqref{eq:final_tilde_p} with ~$w_i^{\star}$.
\STATE Send the new rule's posterior to the FL server.
\ENDFOR
\STATE //At the FL server:
\STATE Receive rule posterior probability from clients.
\STATE Generate samples based on posterior probability.
\STATE Use the generated samples to update $\theta$ by maximizing Eq.~\eqref{eq:max_p_theta} to solve Eq.~\eqref{eq:one_many_FL.a}.
\STATE Distribute new shared prior probabilities $T_{\theta}(r_{\mathrm{head}})$ to each client.
\ENDFOR
\end{algorithmic}
\end{algorithm}

When the data distributions are non-i.i.d., the variational rule distribution $\tilde{p}(\tilde{z})$'s expectation in Eq.~\eqref{eq:vi_proof} in \leref{le.log_to_plus_iid} will differentiate into $\mathbb{E}_{\tilde{p}(\bar{z})}$ on the FL server and $\mathbb{E}_{\tilde{p}(z_{i})}$ on client $i$ for FL personalization purposes. ($\tilde{z}$ is unified symbol for both $\bar{z}$ and $z_{i}$ with i.i.d case.) To ensure the validity of \leref{le.log_to_plus_iid} influenced by the rule distribution heterogeneity, in Eq.~\eqref{eq:one_many_FL_lower}, we add an additional KL-divergence constraint of rule latent variable $D_{\mathrm{KL}}(\tilde{p}_{w_i}(z_i) ||\tilde{p}_{\theta}(\bar{z}))$ to penalize the discrepancy between $\tilde{p}_\theta(\bar{z}$) and $\tilde{p}_{w_{i}}(z_i)$ to reduce the rule distribution heterogeneity. This constraint is added as a regularization term to the Eq.~\eqref{eq:one_many_FL_lower},
where $\lambda$ is the coefficient that balances the distribution distance between the latent variance $z_i$ of client $i$ and the global latent variance $\bar{z}$ of the FL server, and
$\tilde{p}_\theta(\bar{z})$ has been calculated by the server and distributed to the clients.

Substituting $\tilde{p}_{w_i}(z_i)$ of Eq.~\eqref{eq:final_tilde_p} and $\tilde{p}_{\theta}(\bar{z}))$ of Eq.~\eqref{eq:mu_theta} into the definition of KL, $D_{\mathrm{KL}}(\tilde{p}_{w_i}(z_i) ||\tilde{p}_{\theta}(\bar{z}))$ can be re-expressed as:
\vspace{-2mm}
\begin{align}\label{eq:min_w_add_kl_spec} \notag
 D_{\mathrm{KL}}(\tilde{p}_{w_i}(z_i) ||\tilde{p}_{\theta}(\bar{z}))
 &=\tilde{p}_\theta(\bar{z}) \log\left(\frac{\tilde{p}_{w_i}(z_i) }{\tilde{p}_\theta(\bar{z})}\right)
\\
&= \mathrm{Mu}_{\theta}\log\left(\frac{\mathrm{Mu}_{w_i}}{\mathrm{Mu}_{\theta}}\right).\!
\vspace{-2mm}
\end{align}

\vspace{-6mm}
\subsection{Algorithm Design for \methodname{}}
\vspace{-2mm}
In the previous sections, we have provided parameterization for the objective function of the server and each client in Eq.~\eqref{eq:one_many_FL.a}, Eq.~\eqref{eq:one_many_FL.b} and $D_{\mathrm{KL}}(\tilde{p}_{w_i}(z_i) ||\tilde{p}_{\theta}(\bar{z}))$ in Eq~\eqref{eq:one_many_FL_lower} with the KG-based neuro-symbolic scenario.

In this section, a corresponding FL neuro-symbolic algorithm instance \methodname{} is presented in Algorithm~\ref{ag:algorithm}. In Lines 12–15 of the algorithm, the server receives the posterior probability distributions from the clients, creates new posterior rule samples, and incorporates them into the existing pool of rule training samples. Then in server maximizing Eq.~\eqref{eq:max_p_theta} can be converted to minimizing the $\mathcal{L}_{T_{\theta}}$ under the assumption of Eq.\eqref{eq:mu_theta} by utilizing these samples (comprising $r_{\mathrm{body}}$-$r_{\mathrm{head}}$ pairs) to address the subtask of Eq.\eqref{eq:one_many_FL.a}. On the client side with Lines 4–9 of the algorithm, prior distribution probabilities for the candidate rules provided by the server are generated. These candidates are scored using Eq.~\eqref{eq:score_f}. To address the subtask of Eq.\eqref{eq:one_many_FL.b}, the client uses rule scores to build Eq.~\eqref{eq:log_pw_KG} and incorporate the KL-divergence constraint (Eq.~\eqref{eq:min_w_add_kl_spec}) as a regularization term to update the weight $w$. Subsequently, the client uses the updated $w$ to establish a new posterior distribution by Eq.~\eqref{eq:final_tilde_p}. The probabilities of this posterior distribution are then uploaded to the server, marking the commencement of the next round. In Line 5, Line 9, Line 12, and Line 15, the algorithm only transmits the rule prior and posterior probability values between each FL client and the FL server to ensure privacy.

\begin{figure*}[!t]
\centering
\includegraphics[width=0.85\textwidth]{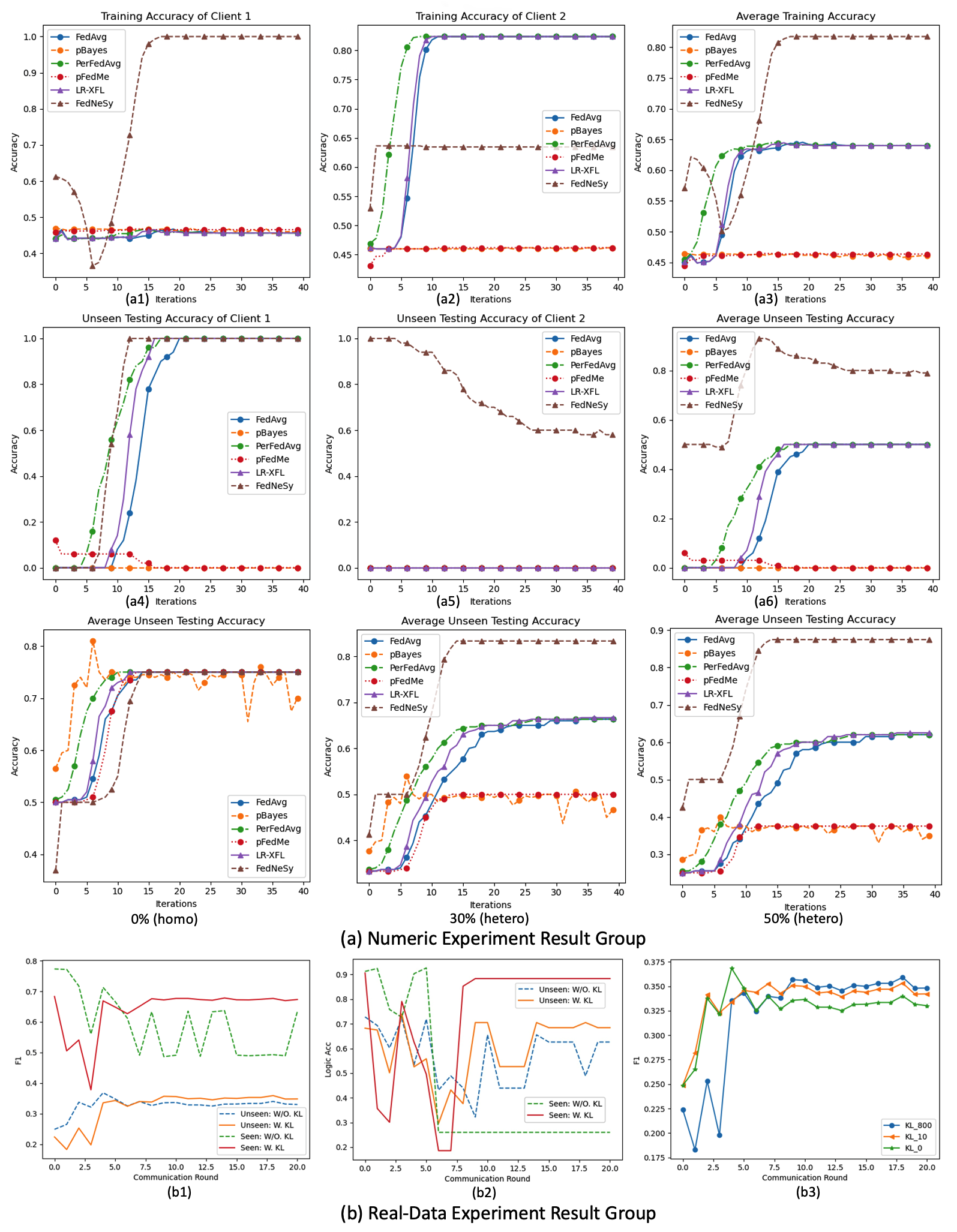}
\caption{Group (a) presents the numerical experiment results. The first row features (a1), (a2) and (a3), which respectively show the training accuracy of the classifiers for client 1, client 2 and the average results. The second row features (a4), (a5) and (a6), which respectively show the unseen testing accuracy for the classifiers of client 1, client 2 and the average results. The third row shows performance comparison results under different ratios of training-testing data heterogeneity: ``0\% (homo)" means training and testing data have the same distribution, while ``33\% (hetero)" and ``50\% (hetero)" indicate that 33\% and 50\% of the unseen testing data, respectively, follow a different distribution from the training data. Group (b) shows the real-data experiment results, including F1-scores in (b1), logic accuracy in (b2) on both the unseen and seen testing data with and without KL-divergence rule distribution constraints (denoted by ``W/O. KL'' and ``W. KL''), and (b3) illustrates how different coefficients of KL-divergence constraint affect the personalization performance.}
\label{fig:exp_overall}
\vspace{-4mm}
\end{figure*}

\vspace{-2mm}
\section{Experimental Evaluation}

\vspace{-1mm}
\subsection{Numerical and Real-Data Experiment Setup}

We conducted two types of experiments: numerical experiments on synthetic data and real-data experiments using the document-level DWIE (Document-Level Web Information Extraction) dataset \cite{zaporojets2021dwie}. For detailed information about the dataset, please refer to Appendix~\ref{appendix:data}.

\vspace{-2mm}
\subsubsection{Numerical Experiment Setup}
In this NSL-based numerical experiment, a federated learning server operating as a three-component Gaussian Mixture Model (3-GMM) is set against two federated learning clients, each equipped with a three-class classifier. The server's 3-GMM is designed to mimic the learning of an overarching distribution of three different rule categories, as depicted in the server part of Figure~\ref{fig:two_flow} (b). Simultaneously, the client-side classifiers are intended to model local rule distributions, each handling two types of rules from the server's set. This setup aims to explore the server-client dynamic in a neuro-symbolic context, where the server learns a global prior rule distribution and the clients focus on partial posterior distributions derived from their classification tasks. The experiment's core goal is to assess whether clients can infer information about an unseen rule distribution through this federated learning process, without direct access to the complete rule set.

In our approach, designated as \methodname{}, we utilize the server's GMM distribution to indirectly facilitate the clients' access to information about the unseen class. This indirect access is made possible through the server's comprehensive modeling of the overall data distribution, which includes the unseen class. For comparative analysis with other methodologies, all baseline methods are detailed in Table~\ref{tb:pfl_comp}. These methods are anticipated to enable access to information about the unseen class through different server objectives, as outlined in Table~\ref{tb:pfl_comp}. This comparison aims to highlight the relative strengths and weaknesses of each approach in a federated learning context, particularly when dealing with limited visibility of the complete data set among the clients.
\vspace{-2mm}
\subsubsection{Real-data Experiment Setup}

Similarly, for the real-data experiment, we design a cross-visible distribution multi-domain testing setup where each client is equipped with three distinct sub-datasets: a seen training set, a seen testing set, and an unseen set. Notably, the distribution of the unseen set differs from that of the seen sets and is excluded from model training. Meanwhile, each client's unseen testing data has the same distribution as the seen training data on the neighboring clients. For this, we partition the $10$ NER categories into two groups, each containing $5$ categories. These categories are subsequently cross-combined to yield $4$ unique category combinations, aligned with the $4$ head-tail combinations for the $4$ clients, respectively. A key aspect of our approach involves ensuring that each client's seen set is misaligned with the unseen set by one category. This means that each client's unseen testing data is the seen training data on the neighboring clients. This strategy results in the creation of $4$ non-i.i.d. datasets, each characterized by different head-tail NER category combinations. A client's seen rule representation can help build other clients' unseen rule representations. It is worth noting that our evaluation setting is consistent with baselines in Table \ref{tb:pfl_comp}. They all similarly set distribution shifts on class labels, and their reasons are the same as ours in setting distribution shifts on the NER category.

\vspace{-2mm}
\subsection{Results and Discussion}




In our numerical experiments, we conducted a detailed comparison of the performance of various PFL baseline models, aiming to exclude the interference of unrelated factors, which often mix in real data samples, as shown in Group (a) of Figure~\ref{fig:exp_overall}. Additionally, we performed ablation studies on a real-world, KG-based dataset. This approach was taken to further assess the impact of upstream rule learning on downstream knowledge graph relationship predictions, as well as its performance on semantic rules, as depicted in Group (b) of Figure~\ref{fig:exp_overall}.
\vspace{-2mm}
\subsubsection{Numerical Experiment Results}

Group (a) of Figure~\ref{fig:exp_overall} shows the performance comparison of various baselines listed in Table~\ref{tb:pfl_comp} on the training set and the unseen test set for the numeric experiment. In the unseen test set with three classes, one class label remains unseen throughout the training, showcasing the information transmission capabilities of different federated server objectives. Specifically, (a1), (a2), and (a3) present the results of individual clients and the average on the training set, while (a4), (a5), and (a6) correspond to the results on the unseen test set. It can be observed that due to unbalanced classes, mishandling heterogeneity can result in high accuracy for some classes and low for others, creating a competing accuracy scenario among clients. This explains why the baseline methods excel in (a2), but perform poorly in (a1). \methodname{} can balance this complementary heterogeneity-induced competing accuracy problem, achieving superior overall average results in both average training accuracy in (a3) and unseen testing accuracy in (a6). We further tested the performance of \methodname{} in comparison with other baselines under different levels of data heterogeneity in the third row of Figure~\ref{fig:exp_overall}. The results show that the higher the degree of heterogeneity, the more advantageous \methodname{} is. Therefore, compared with other methods, \methodname{} addresses the issue of complementary training-testing data heterogeneity across clients more effectively.
\vspace{-2mm}
\subsubsection{Real-data Experiment Results}

We compare the outcomes of two sets of experiments using F1-scores on both the unseen and seen testing data. ``W/O. KL'' and ``W. KL'' denote experiments conducted without and with KL-divergence rule distribution constraints, respectively. The results in Figure~\ref{fig:exp_overall}~(b1) show that introducing a KL-divergence constraint to both the seen and unseen testing data groups leads to convergence at higher F1-scores. Conversely, the unseen testing data group without KL-divergence adjustment achieves convergence but at lower F1-scores. The seen testing data group without KL-divergence constraint exhibits oscillations and does not achieve convergence under the same conditions. Additionally, we analyze the impact of varying KL coefficients on F1 score results for the unseen dataset. The results in Figure~\ref{fig:exp_overall}~(b3) demonstrate that different coefficients indeed influence the personalization performance.

To further assess the logical reasoning capabilities of \methodname{}, we adopt the 39 golden first-order logic predicates from the DWIE dataset \cite{zaporojets2021dwie} for consistency checks after updates by the rule selector model in the lower level, following \cite{ru2021learning}. These predicates include atomic formulas such as $member\_of(X, Y) \wedge sport\_player(X) \rightarrow player\_of(X, Y)$. Logic accuracy is evaluated by plugging predicted relationships from the test set into the rule head and body, respectively. The logic accuracy curves in Figure~\ref{fig:exp_overall}~(b2) correspond to the four groups of experiments mentioned earlier. Consistency with the F1 score results shown in Figure~\ref{fig:exp_overall}~(b1) is evident, with the group exhibiting oscillations achieving the lowest logic accuracy. The groups with KL-divergence constraints achieve higher logic accuracy compared to the groups without such constraints.

\vspace{-3mm}
\section{Conclusions}

In summary, this work introduces a pioneering framework for federated learning, marking the first instance of addressing rule induction heterogeneity and the novel application of distribution-coupled Bilevel Optimization. Our proposed factorizable federated V-EM algorithm effectively manages vast rule search spaces in cross-domain scenarios, significantly boosting computational efficiency. Additionally, our method has demonstrated superior performance in experimental setups making substantial theoretical and practical contributions to the field.


\section*{Acknowledgments}
This research is supported, in part, by the National Research Foundation Singapore and DSO National Laboratories under the AI Singapore Programme (No. AISG2-RP-2020-019); the RIE 2020 Advanced Manufacturing and Engineering (AME) Programmatic Fund (No. A20G8b0102), Singapore; the Joint NTU-WeBank Research Centre on Fintech; and the National Key R\&D Program of China (No. 2021YFF0900800).

\section*{Impact Statements}
This paper introduces advancements in the field of federated neuro-symbolic learning, with a primary focus on the mathematical formulation and discussions related to optimizing model parameters. To the best of our knowledge and understanding, there are no identifiable societal consequences resulting from our work. Hence, we assert that no specific societal impacts need to be highlighted in this context.


\nocite{langley00}

\bibliography{icml2024}

\begin{thebibliography}{39}
\providecommand{\natexlab}[1]{#1}
\providecommand{\url}[1]{\texttt{#1}}
\expandafter\ifx\csname urlstyle\endcsname\relax
  \providecommand{\doi}[1]{doi: #1}\else
  \providecommand{\doi}{doi: \begingroup \urlstyle{rm}\Url}\fi

\bibitem[Achituve et~al.(2021)Achituve, Shamsian, Navon, Chechik, and Fetaya]{achituve2021personalized}
Achituve, I., Shamsian, A., Navon, A., Chechik, G., and Fetaya, E.
\newblock Personalized federated learning with {Gaussian} processes.
\newblock In \emph{Advances in Neural Information Processing Systems}, volume~34, pp.\  8392--8406, 2021.

\bibitem[Bordes et~al.(2013)Bordes, Usunier, Garcia-Duran, Weston, and Yakhnenko]{bordes2013translating}
Bordes, A., Usunier, N., Garcia-Duran, A., Weston, J., and Yakhnenko, O.
\newblock Translating embeddings for modeling multi-relational data.
\newblock In \emph{Advances in Neural Information Processing Systems}, volume~26, 2013.

\bibitem[Ciravegna et~al.(2020)Ciravegna, Giannini, Melacci, Maggini, and Gori]{ciravegna2020constraint}
Ciravegna, G., Giannini, F., Melacci, S., Maggini, M., and Gori, M.
\newblock A constraint-based approach to learning and explanation.
\newblock In \emph{AAAI Conference on Artificial Intelligence}, volume~34, pp.\  3658--3665, 2020.

\bibitem[Ciravegna et~al.(2023)Ciravegna, Barbiero, Giannini, Gori, Li{\'o}, Maggini, and Melacci]{ciravegna2023logic}
Ciravegna, G., Barbiero, P., Giannini, F., Gori, M., Li{\'o}, P., Maggini, M., and Melacci, S.
\newblock Logic explained networks.
\newblock \emph{Artificial Intelligence}, 314:\penalty0 103822, 2023.

\bibitem[Dickens et~al.(2024)Dickens, Gao, Pryor, Wright, and Getoor]{dickens2024convex}
Dickens, C., Gao, C., Pryor, C., Wright, S., and Getoor, L.
\newblock Convex and bilevel optimization for neuro-symbolic inference and learning.
\newblock \emph{arXiv preprint arXiv:2401.09651}, 2024.

\bibitem[Dieuleveut et~al.(2021)Dieuleveut, Fort, Moulines, and Robin]{dieuleveut2021federated}
Dieuleveut, A., Fort, G., Moulines, E., and Robin, G.
\newblock {Federated-EM} with heterogeneity mitigation and variance reduction.
\newblock In \emph{Advances in Neural Information Processing Systems}, volume~34, pp.\  29553--29566, 2021.

\bibitem[Dinh et~al.(2021)Dinh, Vu, Tran, Dao, and Zhang]{dinh2021fedu}
Dinh, C.~T., Vu, T.~T., Tran, N.~H., Dao, M.~N., and Zhang, H.
\newblock Fedu: A unified framework for federated multi-task learning with {Laplacian} regularization.
\newblock \emph{arXiv preprint arXiv:2102.07148}, 400, 2021.

\bibitem[Fallah et~al.(2020{\natexlab{a}})Fallah, Mokhtari, and Ozdaglar]{fallah2020convergence}
Fallah, A., Mokhtari, A., and Ozdaglar, A.
\newblock On the convergence theory of gradient-based model-agnostic meta-learning algorithms.
\newblock In \emph{International Conference on Artificial Intelligence and Statistics}, pp.\  1082--1092. PMLR, 2020{\natexlab{a}}.

\bibitem[Fallah et~al.(2020{\natexlab{b}})Fallah, Mokhtari, and Ozdaglar]{fallah2020personalized}
Fallah, A., Mokhtari, A., and Ozdaglar, A.
\newblock Personalized federated learning with theoretical guarantees: A model-agnostic meta-learning approach.
\newblock In \emph{Advances in Neural Information Processing Systems}, 2020{\natexlab{b}}.

\bibitem[Garcez et~al.(2008)Garcez, Lamb, and Gabbay]{garcez2008neural}
Garcez, A.~S., Lamb, L.~C., and Gabbay, D.~M.
\newblock \emph{Neural-symbolic cognitive reasoning}.
\newblock Springer Science \& Business Media, 2008.

\bibitem[Goebel et~al.(2023)Goebel, Yu, Faltings, Fan, and Xiong]{Goebel-et-al:2023}
Goebel, R., Yu, H., Faltings, B., Fan, L., and Xiong, Z.
\newblock \emph{Trustworthy Federated Learning}, volume 13448.
\newblock Springer, Cham, 2023.

\bibitem[Huang et~al.(2021)Huang, Chu, Zhou, Wang, Liu, Pei, and Zhang]{huang2021personalized}
Huang, Y., Chu, L., Zhou, Z., Wang, L., Liu, J., Pei, J., and Zhang, Y.
\newblock Personalized cross-silo federated learning on non-iid data.
\newblock In \emph{Proceedings of the AAAI Conference on Artificial Intelligence}, volume~35, pp.\  7865--7873, 2021.

\bibitem[Li \& Wang(2019)Li and Wang]{li2019fedmd}
Li, D. and Wang, J.
\newblock Fedmd: Heterogenous federated learning via model distillation.
\newblock \emph{arXiv preprint arXiv:1910.03581}, 2019.

\bibitem[Liu et~al.(2024{\natexlab{a}})Liu, Jiang, Zheng, Chen, Qi, Huang, and Shao]{liu2023bayesian}
Liu, L., Jiang, X., Zheng, F., Chen, H., Qi, G.-J., Huang, H., and Shao, L.
\newblock A {Bayesian} federated learning framework with online laplace approximation.
\newblock \emph{IEEE Transactions on Pattern Analysis and Machine Intelligence}, 2024{\natexlab{a}}.

\bibitem[Liu et~al.(2024{\natexlab{b}})Liu, Xing, Deng, Li, Guan, and Yu]{Liu-et-al:2024TNNLS}
Liu, R., Xing, P., Deng, Z., Li, A., Guan, C., and Yu, H.
\newblock Federated graph neural networks: Overview, techniques and challenges.
\newblock \emph{IEEE Transactions on Neural Networks and Learning Systems}, 2024{\natexlab{b}}.

\bibitem[Louizos et~al.(2021)Louizos, Reisser, Soriaga, and Welling]{louizos2021expectation}
Louizos, C., Reisser, M., Soriaga, J., and Welling, M.
\newblock An expectation-maximization perspective on federated learning.
\newblock \emph{arXiv preprint arXiv:2111.10192}, 2021.

\bibitem[Lu(2023)]{lu2023bilevel}
Lu, S.
\newblock Bilevel optimization with coupled decision-dependent distributions.
\newblock In \emph{International Conference on Machine Learning}, pp.\  22758--22789. PMLR, 2023.

\bibitem[Lu et~al.(2021)Lu, Khan, Akhalwaya, Riegel, Horesh, and Gray]{9415044}
Lu, S., Khan, N., Akhalwaya, I.~Y., Riegel, R., Horesh, L., and Gray, A.
\newblock Training logical neural networks by primal–dual methods for neuro-symbolic reasoning.
\newblock In \emph{IEEE International Conference on Acoustics, Speech and Signal Processing}, pp.\  5559--5563, 2021.

\bibitem[Minervini et~al.(2020)Minervini, Riedel, Stenetorp, Grefenstette, and Rockt{\"a}schel]{minervini2020learning}
Minervini, P., Riedel, S., Stenetorp, P., Grefenstette, E., and Rockt{\"a}schel, T.
\newblock Learning reasoning strategies in end-to-end differentiable proving.
\newblock In \emph{International Conference on Machine Learning}, pp.\  6938--6949. PMLR, 2020.

\bibitem[Nafar et~al.(2023)Nafar, Venable, and Kordjamshidi]{nafar2023teaching}
Nafar, A., Venable, K.~B., and Kordjamshidi, P.
\newblock Teaching probabilistic logical reasoning to transformers.
\newblock \emph{arXiv preprint arXiv:2305.13179}, 2023.

\bibitem[Nickel et~al.(2011)Nickel, Tresp, and Kriegel]{nickel2011three}
Nickel, M., Tresp, V., and Kriegel, H.-P.
\newblock A three-way model for collective learning on multi-relational data.
\newblock In \emph{International Conference on Machine Learning}, pp.\  809–816, 2011.

\bibitem[Nickel et~al.(2012)Nickel, Tresp, and Kriegel]{nickel2012factorizing}
Nickel, M., Tresp, V., and Kriegel, H.-P.
\newblock Factorizing {YAGO}: scalable machine learning for linked data.
\newblock In \emph{International Conference on World Wide Web}, pp.\  271--280, 2012.

\bibitem[Qu et~al.(2020)Qu, Chen, Xhonneux, Bengio, and Tang]{qu2020rnnlogic}
Qu, M., Chen, J., Xhonneux, L.-P., Bengio, Y., and Tang, J.
\newblock Rnnlogic: Learning logic rules for reasoning on knowledge graphs.
\newblock \emph{arXiv preprint arXiv:2010.04029}, 2020.

\bibitem[Rebele et~al.(2016)Rebele, Suchanek, Hoffart, Biega, Kuzey, and Weikum]{rebele2016yago}
Rebele, T., Suchanek, F., Hoffart, J., Biega, J., Kuzey, E., and Weikum, G.
\newblock {YAGO}: A multilingual knowledge base from wikipedia, wordnet, and geonames.
\newblock In \emph{The Semantic Web--ISWC 2016: 15th International Semantic Web Conference}, pp.\  177--185. Springer, 2016.

\bibitem[Riedel et~al.(2013)Riedel, Yao, McCallum, and Marlin]{riedel2013relation}
Riedel, S., Yao, L., McCallum, A., and Marlin, B.~M.
\newblock Relation extraction with matrix factorization and universal schemas.
\newblock In \emph{The 2013 Conference of the North American chapter of the Association for Computational Linguistics: Human Language Technologies}, pp.\  74--84, 2013.

\bibitem[Ru et~al.(2021)Ru, Sun, Feng, Qiu, Zhou, Zhang, Yu, and Li]{ru2021learning}
Ru, D., Sun, C., Feng, J., Qiu, L., Zhou, H., Zhang, W., Yu, Y., and Li, L.
\newblock Learning logic rules for document-level relation extraction.
\newblock \emph{arXiv preprint arXiv:2111.05407}, 2021.

\bibitem[Smith et~al.(2017)Smith, Chiang, Sanjabi, and Talwalkar]{smith2017federated}
Smith, V., Chiang, C.-K., Sanjabi, M., and Talwalkar, A.~S.
\newblock Federated multi-task learning.
\newblock In \emph{Advances in Neural Information Processing Systems}, volume~30, 2017.

\bibitem[T~Dinh et~al.(2020)T~Dinh, Tran, and Nguyen]{t2020personalized}
T~Dinh, C., Tran, N., and Nguyen, J.
\newblock Personalized federated learning with moreau envelopes.
\newblock In \emph{Advances in Neural Information Processing Systems}, volume~33, pp.\  21394--21405, 2020.

\bibitem[Tan et~al.(2021)Tan, Yu, Cui, and Yang]{tan2021towards}
Tan, A.~Z., Yu, H., Cui, L., and Yang, Q.
\newblock Towards personalized federated learning.
\newblock \emph{IEEE Transactions on Neural Networks and Learning Systems}, pp.\  doi:10.1109/TNNLS.2022.3160699, 2021.

\bibitem[Tan et~al.(2022)Tan, Yu, Cui, and Yang]{tan2022towards}
Tan, A.~Z., Yu, H., Cui, L., and Yang, Q.
\newblock Towards personalized federated learning.
\newblock \emph{IEEE Transactions on Neural Networks and Learning Systems}, 34\penalty0 (12):\penalty0 9587--9603, 2022.

\bibitem[Wang et~al.(2014)Wang, Zhang, Feng, and Chen]{wang2014knowledge}
Wang, Z., Zhang, J., Feng, J., and Chen, Z.
\newblock Knowledge graph embedding by translating on hyperplanes.
\newblock In \emph{AAAI Conference on Artificial Intelligence}, pp.\  1112--1119, 2014.

\bibitem[Weston et~al.(2013)Weston, Bordes, Yakhnenko, and Usunier]{weston2013connecting}
Weston, J., Bordes, A., Yakhnenko, O., and Usunier, N.
\newblock Connecting language and knowledge bases with embedding models for relation extraction.
\newblock \emph{arXiv preprint arXiv:1307.7973}, 2013.

\bibitem[Wu et~al.(2020)Wu, He, and Chen]{wu2020personalized}
Wu, Q., He, K., and Chen, X.
\newblock Personalized federated learning for intelligent iot applications: A cloud-edge based framework.
\newblock \emph{IEEE Open Journal of the Computer Society}, 1:\penalty0 35--44, 2020.

\bibitem[Xu et~al.(2022)Xu, Ye, Chen, Zhao, Chen, and Zhang]{xu2022ruleformer}
Xu, Z., Ye, P., Chen, H., Zhao, M., Chen, H., and Zhang, W.
\newblock Ruleformer: Context-aware rule mining over knowledge graph.
\newblock In \emph{International Conference on Computational Linguistics}, pp.\  2551--2560, 2022.

\bibitem[Yang et~al.(2017)Yang, Yang, and Cohen]{yang2017differentiable}
Yang, F., Yang, Z., and Cohen, W.~W.
\newblock Differentiable learning of logical rules for knowledge base completion.
\newblock \emph{arXiv preprint arXiv:1702.08367}, 2017.

\bibitem[Zaporojets et~al.(2021)Zaporojets, Deleu, Develder, and Demeester]{zaporojets2021dwie}
Zaporojets, K., Deleu, J., Develder, C., and Demeester, T.
\newblock Dwie: An entity-centric dataset for multi-task document-level information extraction.
\newblock \emph{Information Processing \& Management}, 58\penalty0 (4):\penalty0 102563, 2021.

\bibitem[Zhang et~al.(2021)Zhang, Chen, Zhang, Ke, and Ding]{zhang2021neural}
Zhang, J., Chen, B., Zhang, L., Ke, X., and Ding, H.
\newblock Neural, symbolic and neural-symbolic reasoning on knowledge graphs.
\newblock \emph{AI Open}, 2:\penalty0 14--35, 2021.

\bibitem[Zhang et~al.(2022)Zhang, Li, Li, Guo, and Shao]{zhang2022personalized}
Zhang, X., Li, Y., Li, W., Guo, K., and Shao, Y.
\newblock Personalized federated learning via variational bayesian inference.
\newblock In \emph{International Conference on Machine Learning}, pp.\  26293--26310. PMLR, 2022.

\bibitem[Zhang \& Yu(2023)Zhang and Yu]{zhang2023lr}
Zhang, Y. and Yu, H.
\newblock Lr-xfl: Logical reasoning-based explainable federated learning.
\newblock \emph{arXiv preprint arXiv:2308.12681}, 2023.

\end{thebibliography}
\bibliographystyle{icml2024}

\newpage
\appendix
\onecolumn
\section{Supplemental Material}

In this material, we provide more detailed discussions on the theory and realization of our \methodname{}.

\subsection{Proof of Lemma \ref{le.log_to_plus_iid}}
\label{appendix:VI}

\begin{proof}
Firstly, in the case of i.i.d. rule distribution between the FL server and clients, we represent the rule latent variable of $z_i$ and $\bar{z}$ using the unified symbol $\tilde{z}$. Hence, it has $\tilde{z}=\tilde{z}_{1:n}$. We define $\tilde{p}(\tilde{z})$ as the variational distribution on the latent rule variable $\tilde{z}$. Consequently, we can obtain
\begin{align} \label{eq:vi_proof} \notag
&\quad\;\log \left( p_{w_{1:n},\theta}(a_{1:n}|q_{1:n},\mathcal{G}_{1:n}) \right) \\\notag
&= \int \tilde{p}(\tilde{z}) \log \left( p_{w_{1:n},\theta}(a_{1:n}|q_{1:n},\mathcal{G}_{1:n}) \right) d\tilde{z} \\\notag
&= \int \tilde{p}(\tilde{z}) \log \left( \frac{p_{w_{1:n},\theta}(a_{1:n}|q_{1:n},\mathcal{G}_{1:n})p_{w_{1:n},\theta}(\tilde{z}| a_{1:n},q_{1:n},\mathcal{G}_{1:n})}{p_{w_{1:n},\theta}(\tilde{z}| a_{1:n},q_{1:n},\mathcal{G}_{1:n})} \right) d\tilde{z} \\\notag
&= \int \tilde{p}(\tilde{z}) \log \left( \frac{p_{w_{1:n},\theta}(a_{1:n},\tilde{z}| q_{1:n},\mathcal{G}_{1:n})}{p_{w_{1:n},\theta}(\tilde{z}| a_{1:n},q_{1:n},\mathcal{G}_{1:n})} \right) d\tilde{z} \\\notag
&= \int \tilde{p}(\tilde{z}) \log \left( \frac{p_{w_{1:n},\theta}(a_{1:n},\tilde{z}| q_{1:n},\mathcal{G}_{1:n})\tilde{p}(\tilde{z})}{p_{w_{1:n},\theta}(\tilde{z}| a_{1:n},q_{1:n},\mathcal{G}_{1:n})\tilde{p}(\tilde{z})} \right) d\tilde{z} \\\notag
&= \int \tilde{p}(\tilde{z}) \log \left( \frac{p_{w_{1:n},\theta}(a_{1:n},\tilde{z}| q_{1:n},\mathcal{G}_{1:n})}{\tilde{p}(\tilde{z})} \right) d\tilde{z} - \int \tilde{p}(\tilde{z}) \log \left( \frac{p_{w_{1:n},\theta}(\tilde{z}| a_{1:n},q_{1:n},\mathcal{G}_{1:n})}{\tilde{p}(\tilde{z})} \right) d\tilde{z} \\ \notag
&= \mathbb{E}_{\tilde{p}(\tilde{z})} \log \left( \frac{p_{w_{1:n},\theta}(a_{1:n},\tilde{z}| q_{1:n},\mathcal{G}_{1:n})}{\tilde{p}(\tilde{z})} \right) - \mathbb{E}_{\tilde{p}(\tilde{z})} \log \left( \frac{p_{w_{1:n},\theta}(\tilde{z}| a_{1:n},q_{1:n},\mathcal{G}_{1:n})}{\tilde{p}(\tilde{z})} \right)
\\
&= \mathbb{E}_{\tilde{p}(\tilde{z})} \log \left( \frac{p_{w_{1:n},\theta}(a_{1:n},\tilde{z}| q_{1:n},\mathcal{G}_{1:n})}{\tilde{p}(\tilde{z})} \right) - \sum_{i=1}^{n}\mathbb{E}_{\tilde{p}(\tilde{z})} \log \left( \frac{p_{w_{i},\theta}(\tilde{z}| a_{i},q_{i},\mathcal{G}_{i})}{\tilde{p}(\tilde{z})} \right).
\end{align}

Following this, the term $\mathbb{E}_{\tilde{p}(\tilde{z})}  \log\left( \frac{p_{w_{1:n},\theta}(a_{1:n},\tilde{z}| q_{1:n},\mathcal{G}_{1:n})}{\tilde{p}(\tilde{z})}\right)$ in Eq~\eqref{eq:vi_proof} is defined as $\mathcal{L}_{\mathrm{ELBO}}(\tilde{p}(\tilde{z}),p_{w_{1:n},\theta}(\tilde{z}))$, which is the evidence lower bound (ELBO) of the log-likelihood function. Additionally, the term $- \mathbb{E}_{\tilde{p}(\tilde{z})}  \log\left( \frac{p_{w_{i},\theta}(\tilde{z}| a_{i},q_{i},\mathcal{G}_{i})}{\tilde{p}(\tilde{z})}\right)$ aligns with the definition of KL-divergence, denoted by $D_{\mathrm{KL}}(\tilde{p}(\tilde{z})||p_{w_{i},\theta}(\tilde{z}|q_{i},a_{i},\mathcal{G}_{i}))$. Thus, we can rewrite Eq~\eqref{eq:vi_proof} as:
\begin{align} \label{eq:vi_temp} \notag
&\quad\;\log \left( p_{w_{1:n},\theta}(a_{1:n}|q_{1:n},\mathcal{G}_{1:n}) \right) \\
&=\mathcal{L}_{\mathrm{ELBO}}(\tilde{p}(\tilde{z}),p_{w_{1:n},\theta}(\tilde{z}))+\sum_{i=1}^{n}D_{\mathrm{KL}}(\tilde{p}(\tilde{z})||p_{w_{i},\theta}(\tilde{z}|q_{i},a_{i},\mathcal{G}_{i})).
\end{align}

Since $\forall i$, KL-divergence $D_{\mathrm{KL}}(\tilde{p}(\tilde{z})||p_{w_{i},\theta}(\tilde{z}|q_{i},a_{i},\mathcal{G}_{i}))$ is non-negative, so the $\sum_{i=1}^{n}D_{\mathrm{KL}}(\tilde{p}(\tilde{z})||p_{w_{i},\theta}(\tilde{z}|q_{i},a_{i},\mathcal{G}_{i}))$ is non-negative. the ELBO $\mathcal{L}_{\mathrm{ELBO}}(\tilde{p}(\tilde{z}),p_{w_{1:n},\theta}(\tilde{z}))$ is maximized when $\sum_{i=1}^{n}D_{\mathrm{KL}}(\tilde{p}(\tilde{z})||p_{w_{i},\theta}(\tilde{z}|q_{i},a_{i},\mathcal{G}_{i}))=0$. Additionally, the $\sum_{i=1}^{n}D_{\mathrm{KL}}(\tilde{p}(\tilde{z})||p_{w_{i},\theta}(\tilde{z}|q_{i},a_{i},\mathcal{G}_{i}))$ can be factored into each client $i$. Therefore, considering the i.i.d. case where $\bar{z}=z_i=\tilde{z}$, for FL setting, maximizing the overall log-likelihood function is achieved by maximizing the shared $\mathcal{L}_{\mathrm{ELBO}}(\tilde{p}(\bar{z}),p_{\theta,w_{1:n}}(\bar{z}))$ on FL server and minimizing $D_{\mathrm{KL}}(\tilde{p}(z_i)||p_{w_{i},\theta}(z_i|q_{i},a_{i},\mathcal{G}_{i}))$ on each client $i$, and the Eq~\eqref{eq:vi_temp} can be rewritten as Eq~\eqref{eq:vi_iid}.

\end{proof}

\subsection{Proof of Lemma \ref{le.mstep}}

\begin{proof}
Following the \leref{le.log_to_plus_iid}, we can maximize $\mathcal{L}_{\mathrm{ELBO}}(\tilde{p}(\tilde{z}),p_{w_{1:n},\theta}(\tilde{z}))$ on the FL server using the M-step of the V-EM algorithm by updating the decision weight $\theta$. The term $\mathcal{L}_{\mathrm{ELBO}}$ is further rewritten as:
\begin{align} \label{eq:elbo} \notag
&\quad\; \mathcal{L}_{\mathrm{ELBO}}(\tilde{p}(\tilde{z}),p_{w_{1:n},\theta}(\tilde{z}))
\\ \notag
&=\mathbb{E}_{\tilde{p}(\tilde{z})}  \log\left( \frac{p_{w_{1:n},\theta}(a_{1:n},\tilde{z}| q_{1:n},\mathcal{G}_{1:n})}{\tilde{p}(\tilde{z})}\right)
\\ 
&= \mathbb{E}_{\tilde{p}(\tilde{z})}\log p_{w_{1:n}}(a_{1:n}|\tilde{z},q_{1:n},\mathcal{G}_{1:n})+\mathbb{E}_{\tilde{p}(\tilde{z})}\log p_\theta(\tilde{z}))-\mathbb{E}_{\tilde{p}(\tilde{z})}\log  \tilde{p}(\tilde{z}).
\end{align}

There is only one term, $\mathbb{E}_{\tilde{p}(\tilde{z})}\log\;p_\theta(\tilde{z})$, that is relevant to $p_{\theta}$. Therefore,
$\max_{w_{1:n},\theta}\mathcal{L}_{\mathrm{ELBO}}(\tilde{p}(\tilde{z}),p_{w_{1:n},\theta}(\tilde{z}))$ can be converted into $
\max_{\theta}\mathbb{E}_{\tilde{p}(\tilde{z})}\log\;p_\theta(\tilde{z})$.

\end{proof}

\subsection{Proof of Lemma \ref{le.estep}}
\begin{proof}
Following the \leref{le.log_to_plus_iid}, the E-step on each client $i$ is designed to minimize $D_{\mathrm{KL}}(\tilde{p}(\tilde{z})||p_{w_{i},\theta}(\tilde{z}|q_{i},a_{i},\mathcal{G}_{i}))$.
The objective of the E-step on each client $i$ is to update $\tilde{p}_{w_i,\theta}(\tilde{z})$, which can be formalized as:
\vspace{-2mm}
\begin{align}\label{eq:KLqp}
\min_{\tilde{p}_{w_i,\theta}(\tilde{z})} D_{\mathrm{KL}}(\tilde{p}_{w_i,\theta}(\tilde{z})||p_{w_i,\theta}(\tilde{z}|q_i,a_i,\mathcal{G}_i)).
\vspace{-2mm}
\end{align}

Then, we re-write Eq. \eqref{eq:KLqp} as a variational distribution expectation form ($\mathbb{E}_{\tilde{p}(\tilde{z})}$) to transform the objective of finding a probability density function  (PDF) for $\min_{\tilde{p}_{w_i,\theta}(\tilde{z})}$ into finding a solution weight for $\max_{w_i,\theta}$ as follows,
\vspace{-2mm}
\begin{align}\label{eq:tilde_fi} \notag
&\quad\;\max_{w_i,\theta}\mathbb{E}_{\tilde{p}(\tilde{z})}\log(p_{w_i,\theta}(\tilde{z}|q_i,a_i,\mathcal{G}_i))
 \\ \notag
& = \max_{w_i,\theta}\mathbb{E}_{\tilde{p}(\tilde{z})}\log(p_{w_i}(a_i|\tilde{z},q_i,\mathcal{G}_i) 
  p_\theta(\tilde{z})) 
 \\
&=\max_{w_i,\theta}\mathbb{E}_{\tilde{p}(\tilde{z})}(\log p_{w_i}(a_i|\tilde{z},q_i,\mathcal{G}_i))   + \mathbb{E}_{\tilde{p}(\tilde{z})}(\log p_\theta(\tilde{z})).
\end{align}
From Eq.~\eqref{eq:tilde_fi}, it can be observed that the term  $\mathbb{E}_{\tilde{p}(\tilde{z})}(\log p_\theta(\tilde{z})$ has been fixed in the M-step and has no relationship with $w_i$. 
$\mathbb{E}_{\tilde{p}(\tilde{z})}(\log p_{w_i}(a_i|\tilde{z},q_i,\mathcal{G}_i))$ is key for solving Eq. \eqref{eq:KLqp} and it is a function in terms of $w_i$.
Therefore, the lower-level optimization problem on the client $i$ can be formalized as
\vspace{-2mm}
\begin{align}
\max_{w_i} \mathbb{E}_{\tilde{p}(\tilde{z})}\left[\log p_{w_i}(a_i|\tilde{z},q_i,\mathcal{G}_i)\right].
\end{align}
\end{proof}

\subsection{Proof of Lemma \ref{le.tilde_p}}

\begin{proof}
Following the \leref{le.estep}, after we get lower level solution weight $w^{\star}$ to meet Eq.~\eqref{eq:KLqp}, we can also use $w^{\star}$ to calculate the approximated posterior for uploading to server for next round of V-EM.
For that, we re-write $\tilde{p}_{w_i,\theta}(\tilde{z})$ in the following log-form, i.e.,
\begin{align}\label{eq:new_pp} \notag
\tilde{p}_{w_i,\theta}(\tilde{z})
 &\propto  \exp(\log(\mathbb{E}_{\tilde{p}(\tilde{z})}p_{w_i,\theta}(\tilde{z}|q_i,a_i,\mathcal{G}_i)) )
 \\ \notag
& \propto\exp(\log(\mathbb{E}_{\tilde{p}(\tilde{z})}p_{w_i}(a_i|\tilde{z},q_i,\mathcal{G}_i) 
  p_\theta(\tilde{z})) )
 \\
&\propto\exp((\log(\mathbb{E}_{\tilde{p}(\tilde{z})} p_{w_i}(a_i|\tilde{z},q_i,\mathcal{G}_i))   +(\log (\mathbb{E}_{\tilde{p}(\tilde{z})}p_\theta(\tilde{z})))).
\end{align}


Since we can get $\mathbb{E}_{\tilde{p}(\tilde{z})} \log( p_{w_i}(a_i|\tilde{z},q_i,\mathcal{G}_i))$ with Eq.~\eqref{eq:log_pw_KG} and get $\mathbb{E}_{\tilde{p}(\tilde{z})}\log(p_\theta(\tilde{z}))$ from Eq.~\eqref{eq:mu_theta}, we can get the approximated posterior as follows:

\begin{align} \label{eq:H_to_mu} \notag
&\tilde{p}_{w_i,\theta}(\tilde{z})
\\ \notag
&\propto \exp\left( \frac{1}{2}y(r_{\mathrm{head}}) \cdot \sum_{\substack{j=1 \\ w_{ij} \in w_i \\z_{ij} \in \tilde{z}}}^{J} w_{ij} \cdot \max_{\mathcal{G}_i}{\prod_{\substack{k=1 \\ r_1\wedge ...\wedge r_l \in z_{ij}}}^{l}x(r_{kj})} +\log(\mathrm{Mu}_{\theta}(\tilde{z}|N,T_{\theta}(r_1\wedge ...\wedge r_l|r_{\mathrm{head}}))) \right)
\\ \notag
&\propto \exp\left( \frac{1}{2}y(r_{\mathrm{head}}) \cdot \sum_{\substack{j=1 \\ w_{ij} \in w_i \\z_{ij} \in \tilde{z}}}^{J} w_{ij} \cdot \max_{\mathcal{G}_i}\!\!\!\!\!{\prod_{\substack{k=1 \\ r_1\wedge ...\wedge r_l \in z_{ij}}}^{l}\!\!\!\!\!\!x(r_{kj})} +\log\left(\frac{N!}{\prod_{j=1}^{J}n_{j}!}\prod_{j=1}^{J}T_{\theta}(r_1\wedge ...\wedge r_l|r_{\mathrm{head}})\right) \right)
\\ \notag
&\propto \exp\left( \frac{1}{2}y(r_{\mathrm{head}}) \cdot \sum_{\substack{j=1 \\ w_{ij} \in w_i \\z_{ij} \in \tilde{z}}}^{J} w_{ij} \cdot \max_{\mathcal{G}_i}\!\!\!{\prod_{\substack{k=1 \\ r_1\wedge ...\wedge r_l \in z_{ij}}}^{l}\!\!\!\!\!\!x(r_{kj})} +\log\left(\frac{N!}{\prod_{j=1}^{J}n_{j}!}\right)+\log\left(\prod_{j=1}^{J}T_{\theta}(r_1\wedge ...\wedge r_l|r_{\mathrm{head}})\right) \right)
\\ \notag
&\propto \exp\left( \frac{1}{2}y(r_{\mathrm{head}}) \cdot \sum_{\substack{j=1 \\ w_{ij} \in w_i \\z_{ij} \in \tilde{z}}}^{J} w_{ij} \cdot \max_{\mathcal{G}_i}\!\!\!\!{\prod_{\substack{k=1 \\ r_1\wedge ...\wedge r_l \in z_{ij}}}^{l}\!\!\!\!\!\!\!x(r_{kj})}+\log\left(\frac{N!}{\prod_{j=1}^{J}n_{j}!}\right)+\sum_{j=1}^{J}\log(T_{\theta}(r_1\wedge ...\wedge r_l|r_{\mathrm{head}})) \right)
\\ 
&\propto \exp\left( \sum_{j=1}^{J} \left(\frac{1}{J}\cdot \frac{1}{2}y(r_{\mathrm{head}}) \cdot \sum_{\substack{j=1 \\ w_{ij} \in w_i \\z_{ij} \in \tilde{z}}}^{J} w_{ij} \cdot \max_{\mathcal{G}_i}\!\!\!\!\!\!\!{\prod_{\substack{k=1 \\ r_1\wedge ...\wedge r_l \in z_{ij}}}^{l}\!\!\!\!\!\!\!\!x(r_{kj})}+\log(T_{\theta}(r_1\wedge ...\wedge r_l|r_{\mathrm{head}}))\right) +\log\left(\frac{N!}{\prod_{j=1}^{J}n_{j}!}\right) \right),
\end{align}
where $n_{j}$ is the number of times a rule appears in the set $\tilde{z}$.

According to the definition of $\mathcal{H}_{w_i}(\cdot)$ in Eq~\eqref{eq:score_f}, we can rewrite Eq.~\eqref{eq:H_to_mu} as follows:
\begin{align} \label{eq:H_to_mu_2} \notag
&\propto \exp\left(\sum_{j=1}^{J} \mathcal{H}_{w_i}(r_1\wedge ...\wedge r_l|r_{\mathrm{head}})+\log\left(\frac{N!}{\prod_{j=1}^{J}n_{j}!}\right) \right)
\\ \notag
&\propto\exp\left(\log\left(\frac{N!}{\prod_{j=1}^{J}n_{j}!}\right)\right) \exp\left(\sum_{j=1}^{J} \mathcal{H}_{w_i}(r_1\wedge ...\wedge r_l|r_{\mathrm{head}}) \right)
\\ \notag
&\propto\frac{N!}{\prod_{j=1}^{J}n_{j}!} \prod_{j=1}^{J}\exp\left( \mathcal{H}_{w_i}(r_1\wedge ...\wedge r_l|r_{\mathrm{head}}) \right)
\\
&\propto \mathrm{Mu}_{w_i}\left(\tilde{z}|N,\prod_{j=1}^{J}\exp\left( \mathcal{H}_{w_i}(r_1\wedge ...\wedge r_l|r_{\mathrm{head}})\right)\right).
\end{align}

\end{proof}

\subsection{Dataset}
\label{appendix:data}


In our numerical experiment, we synthesized a dataset by generating 600 two-dimensional data points across three classes, each defined by a distinct Gaussian distribution with specific means and covariances: Class 0 around \([20, 20]\) with covariance \(\begin{bmatrix}0.5 & 0.1\\ 0.1 & 0.5\end{bmatrix}\), Class 1 at \([10, 10]\) with \(\begin{bmatrix}0.5 & 0.2\\ 0.2 & 0.5\end{bmatrix}\), and Class 2 positioned at \([2, 2]\) with \(\begin{bmatrix}0.5 & -0.3\\ -0.3 & 0.5\end{bmatrix}\). Labels [0, 1, 2] were assigned to these distributions. For data division, the first client's training set included points from Classes 0 and 1, and 150 points from Class 2 were randomly relabeled with [0, 1, 2], using the rest as the test set. The second client's training set consisted of Classes 1 and 2 points, with additional points from Class 0 randomly labeled for training and some reserved for testing.

For the real-data experiment, we utilize a document-level DWIE (Document-Level Web Information Extraction) dataset \cite{zaporojets2021dwie} that has been pre-processed following the methods outlined in \cite{ru2021learning}. Table~\ref{tb:logic_predicate} encompasses all the first-order logic rule predicates involved in the dataset. This dataset contains detailed word type statistics, particularly named entity recognition (NER) categories, for all selected entity words within news articles. It encompasses $10$ different NER categories. The dataset also features annotated relationships between entities, spanning a total of $65$ relationship categories. We leverage these relation annotations among entities to formulate a relation prediction task. The dataset comprises $799$ documents in total, categorized into $10$ NER types. These documents are distributed across $4$ FL clients, with each client containing $200$ documents, except for the last client, which holds $199$ documents. Within each client, the documents are further divided into training and testing subsets, maintaining a $3:1$ ratio.

\subsection{Model Setup}

For numeric experiment, we use an integrated model setup combining deep learning classifiers with a GMM tailored for a federated learning context. The setup features two neural network classifiers, each with an input dimension of 2 to accommodate the two-dimensional features of our synthetic dataset, a hidden layer comprising 64 units to capture complex data patterns without overfitting, and an output layer with 3 units equipped with a softmax function for 3-class classification. Parallelly, the GMM is configured with 3 components to correspond with the dataset's 3 classes, where the means are initialized based on preliminary data analysis or classifier outputs, and covariance matrices are set to reflect initial data variance, facilitating adaptive learning through the EM processing. 

For the real-data experiment, a transformer model at the server encodes NER categories and relations through unique numerical identifiers. The embedding layer size for relations is $(256, 2R+1)$, with $R$ representing the number of relations (65 in this case). Similarly, the NER category embedding layer has dimensions $(256, 10)$, reflecting the 10 distinct NER categories. The model contains two encoding and decoding layers, and an output layer of size $(256, 2R+1)$. The input for the transformer is obtained by concatenating a rule head and a rule body, both of size 4. Any shortfall is supplemented with padding symbols and masked using a $4\times4$ position mask. Each rule head generates a set of 50 rule bodies on average, and these candidates are then transmitted to the lower-level model, with duplicates removed. On the client side, the rule selector model dynamically initializes the weights for each communication round. These weights are tailored to the candidate rule bodies originating from the upper level, with their variability stemming from the stochastic nature of the upper-level optimization process and the probabilistic characteristics of the generated rules. The weight group size is denoted as $(K, J\times1)$, where $K$ is the number of rule head categories. Each weight group member corresponds to a candidate rule body, with its dimensions as $J\times1$.

\begin{table}[h]
\centering
\begin{tabular}{ll}
 $ \neg spouse\_of \rightarrow spouse\_of$ & 
 $ \neg vs \rightarrow vs$ \\ 
 $ won\_vs \rightarrow vs$ &  
 $ \neg won\_vs \rightarrow vs$ \\ 
 $ \neg child\_of \rightarrow parent\_of$ & 
 $ \neg parent\_of \rightarrow child\_of$ \\ 
 $ ministry\_of \rightarrow agency\_of$ & 
 $ agency\_of\mbox{-}x \wedge  gpe0 \rightarrow agency\_of$ \\ 
 $ agency\_of \wedge  \neg gpe0 \rightarrow agency\_of\mbox{-}x$ & 
 $ agent\_of\mbox{-}x \wedge  gpe0 \rightarrow agent\_of$ \\ 
 $ agent\_of \wedge  \neg gpe0 \rightarrow agent\_of\mbox{-}x$ & 
 $ minister\_of \rightarrow agent\_of$ \\ 
 $ head\_of\_gov \rightarrow agent\_of$ & 
 $ head\_of\_state \rightarrow agent\_of$ \\ 
 $ citizen\_of\mbox{-}x \wedge  gpe0 \rightarrow citizen\_of$ & 
 $ citizen\_of \wedge  \neg gpe0 \rightarrow citizen\_of\mbox{-}x$ \\ 
 $ minister\_of\mbox{-}x \wedge  gpe0 \rightarrow minister\_of$ & 
 $ minister\_of \wedge  \neg gpe0 \rightarrow minister\_of\mbox{-}x$ \\ 
 $ head\_of\_state\mbox{-}x \wedge  gpe0 \rightarrow head\_of\_state$ & 
 $ head\_of\_state \wedge  \neg gpe0 \rightarrow head\_of\_state\mbox{-}x$ \\ 
 $ head\_of\_gov\mbox{-}x \wedge  gpe0 \rightarrow head\_of\_gov$ & 
 $ head\_of\_gov \wedge  \neg gpe0 \rightarrow head\_of\_gov\mbox{-}x$ \\ 
 $ in0\mbox{-}x \wedge  gpe0 \rightarrow in0$ & 
 $ in0 \wedge  \neg gpe0 \rightarrow in0\mbox{-}x$ \\ 
 $ in2 \wedge  in0 \rightarrow in0$ & 
 $ in1 \wedge  in0 \rightarrow in0$ \\ 
 $ based\_in2 \wedge  in0 \rightarrow based\_in0$ & 
 $ based\_in1 \wedge  in0 \rightarrow based\_in0$ \\ 
 $ event\_in2 \wedge  in0 \rightarrow event\_in0$ & 
 $ event\_in1 \wedge  in0 \rightarrow event\_in0$ \\ 
 $ head\_of \rightarrow member\_of$ & 
 $ coach\_of \rightarrow member\_of$ \\ 
 $ spokesperson\_of \rightarrow member\_of$ & 
 $ mayor\_of \rightarrow head\_of\_gov$ \\ 
 $ directed\_by \rightarrow created\_by$ & 
 $ \neg played\_by \wedge  character\_in \rightarrow plays\_in$ \\ 
 $ institution\_of \rightarrow part\_of$ & 
 $ based\_in0\mbox{-}x \wedge  gpe0 \rightarrow based\_in0$ \\ 
 $ based\_in0 \wedge  \neg gpe0 \rightarrow based\_in0\mbox{-}x$ 
 \end{tabular}
\caption{First-order logic predicate using in evaluation}
\label{tb:logic_predicate}
\end{table}

\subsection{Related Works}
\label{appendix:related}

\subsubsection{Comparison of Existing Personalization Federated Learning} 

We additionally analyze current personalized federated learning (PFL) paradigms to demonstrate the necessity for a new PFL approach, one that more effectively addresses the requirements of federated learning for neuro-symbolic learning. Each of these paradigms offers a different strategy for integrating personalization into Federated Learning, aiming to balance the benefits of a global model with the specific needs of individual clients:

\begin{itemize}

\item Regularization-Based PFL 

The formulation $\min_{\{w_k\}} \sum_{k=1}^{K} \left( F_k(w_k) + \lambda \|w_k - w_g\|^2 \right)$ in regularization-based PFL is designed to find a balance between local model accuracy and global model consistency. In this setting, each client $k$ works on optimizing its own model parameters $w_k$, guided by its local loss function $F_k(w_k)$. The regularization term $ \lambda \|w_k - w_g\|^2 $ acts as a bridge, tying the local models to the global model parameters $ w_g $. The regularization coefficient $ \lambda $ is crucial here; it controls how closely each local model adheres to the global model, ensuring that while each model is personalized for local data, it doesn't diverge significantly from the shared global insights. This formulation can cover a range of methods, including FedU \cite{dinh2021fedu}, pFedMe \cite{t2020personalized}, FedAMP \cite{huang2021personalized}.

\item Meta-Learning Based PFL

Meta-learning \cite{fallah2020convergence} in the federated setting, represented by the two-step process of $ w' = w - \beta \nabla_w \sum_{k=1}^{K} F_k(w) $ followed by $ w_k = w' - \alpha \nabla_{w'} F_k(w') $ for each client $ k $, is about learning a model that can quickly adapt to new environments or data distributions. The initial step adjusts the global model parameters $ w $ using a learning rate $ \beta $ and the aggregated loss from all clients. This forms an updated global model $ w' $. Then, in a crucial personalization step, each client fine-tunes this model to their local dataset. The local adaptation uses another learning rate $ \alpha $, allowing each client to adjust the model $ w' $ to better fit their specific data characteristics, resulting in a personalized model $ w_k $. The Per-FedAvg \cite{fallah2020personalized} represents pioneering research work among meta-learning based PFL works.

\item Multi-Task Based PFL

In multi-task based PFL e.g., \cite{smith2017federated,wu2020personalized,li2019fedmd}, encapsulated by problem $ \min_{w_g, \{w_k\}} \sum_{k=1}^{K} F_k(w_g, w_k) $, the learning process is akin to handling multiple related tasks simultaneously. Here, $ w_g $ denotes the shared global parameters that capture commonalities across all clients, while $ \{w_k\} $ represents a collection of client-specific parameters, allowing each client to address its unique aspects. The loss function $ F_k(w_g, w_k) $ for each client is influenced by both of these sets of parameters. This hybrid parameter system enables the model to learn general patterns through the global parameters while also catering to specific client requirements through the local parameters.

\item EM-Based PFL 

The EM-based approach in PFL (e.g., FedSparse \cite{louizos2021expectation} and FedEM \cite{dieuleveut2021federated}) is characterized by a cyclic process of local and global updates. The local updates (E-step) involve each client working with their data and the current global model to estimate local parameters or latent variables. The global update (M-step) then synthesizes these local estimates to refine the global model. This iterative process, while not represented by a single formula, effectively combines the benefits of personalized models with the strength of a globally consistent framework. The EM cycle ensures that each client's model is individually tailored, while the global model continuously integrates these individual learnings, maintaining a coherent overall structure.

\item Bayesian-Based PFL 

Bayesian methods in PFL, described by $ P(w_k|D_k, w_g) \propto P(D_k|w_k)P(w_k|w_g) $, offer a probabilistic approach to model personalization. In this framework, $ P(w_k|D_k, w_g) $ represents the posterior distribution of the parameters for each client $ k $, given their local data $ D_k $ and the global parameters $ w_g $. This approach combines the likelihood of observing the local data under the given parameters ($ P(D_k|w_k) $) with a prior distribution that ties the local parameters to the global model ($ P(w_k|w_g) $). This probabilistic blending allows each client's model to be personalized based on their data while being informed and constrained by the broader insights of the global model. In related works, their algorithmic instances are slightly different.
pFedGP \cite{achituve2021personalized} employs a Gaussian process tree, while pFedBayes \cite{zhang2022personalized} utilizes a Bayesian Neural Network (BNN). FLOA \cite{liu2023bayesian} uses Laplace approximation in Bayesian theory to interpret FL heterogeneity, realizing the discussed paradigm.

\end{itemize}

However, the mentioned above PFL frameworks fail to deal with distribution-coupled federated NSL between server and local levels. The reasons are twofold: first, except for EM-based PFL methods like FedEM, others can't handle PFL involving hidden variables. Second, even in existing EM-based PFLs, such as FedEM, the weights learned are only relevant to local tasks and do not involve additional weights for hidden variables, meaning they can't generate inductive samples. Hence, a new framework capable of handling nested learning of local weights and weights for hidden variables is needed.

\subsection{Further Optimizing Client-Side Computational Complexity}

We have optimized server-side computational complexity by learning a generative rule distribution to sample the unseen rule to reduce the computational complexity. In this part, we further strive to enhance our algorithm by refining the client-side optimization process of the score function to reduce client-side computational complexity.

Recall that in Algorithm \ref{ag:algorithm}, at the lower level, the candidate rule fuzzy value is calculated in local objective function Eq.\eqref{eq:log_pw_KG} and score function Eq.\eqref{eq:score_f}. These functions calculate the fuzzy value for candidate rules by maximizing across the entirety of $\mathcal{G}_i$. Given the time-intensive nature of this step, optimizing the fuzzy value calculation becomes necessary.

The core concept of an improved version of algorithm~\ref{ag:algorithm2} revolves around utilizing original fuzzy value calculation method solely for computing the posterior during the initial $I$ communication rounds, aiming to establish a fundamental rule generator. In subsequent stages, an innovatively designed path-based score function is adopted, replacing the graph-based score function for saving computational time.

Te be specifically, we use ${\max_{\substack{k=1 \\ r_1\wedge ...\wedge r_l \in z_{ij}}}^{l}x(r_{kj})}$ as a replacement for $\max_{\mathcal{G}_i}{\prod_{\substack{k=1 \\ r_1\wedge ...\wedge r_l \in z_{ij}}}^{l}x(r_{kj})}$ in Eq.\eqref{eq:log_pw_KG} and Eq.\eqref{eq:score_f}. In the latter expression, which is a graph-based score function, $\prod_{k \in l}x(r_{kj})$ represents the multiplication of score values along a given path, and $\max_{\mathcal{G}_i}$ indicates finding the shortest path across $\{\mathcal{G}_i\}$. This operation of finding the shortest path is time-consuming. In the former expression, the process is simplified by directly adopting the maximum value among the corresponding score values along the trajectory of the rule body. This serves as a comprehensive score for the rule body in the novel path-based score function.

After this replacement, the new local objective function can be written as follows:

\vspace{-2mm}
\begin{align}\label{eq:log_pw_KG_new}\notag
    &\quad\;\mathbb{E}_{\tilde{p}(z_i)} \log \;p_{w_i}(a_i|z_i,q_i,\mathcal{G}_i)
    \\
&\approx \frac{1}{2}y(r_{\mathrm{head}})\cdot\sum_{\substack{j=1 \\ w_{ij} \in w_i \\z_{ij} \in z_i}}^{J} w_{ij} \cdot {\max_{\substack{k=1 \\ r_1\wedge ...\wedge r_l \in z_{ij}}}^{l}x(r_{kj})}.
    \vspace{-4mm}
\end{align}

And the new score function $\tilde{\mathcal{H}}_{w_i}$ can be written as follows:
\begin{align} \label{eq:score_f_new} \notag
    &\tilde{\mathcal{H}}_{w_i}(r_1\wedge ...\wedge r_l|r_{\mathrm{head}})
    \\ 
    &=\frac{1}{J}\cdot \frac{1}{2}y(r_{\mathrm{head}})\cdot \sum_{\substack{j=1 \\ w_{ij} \in w_i \\z_{ij} \in z_i}}^{J} w_{ij} \cdot {\max_{\substack{k=1 \\ r_1\wedge ...\wedge r_l \in z_{ij}}}^{l}x(r_{kj})}
    + \log(T_{\theta}(r_1\wedge ...\wedge r_l|r_{\mathrm{head}})).
\end{align}

The new approximated posterior $\tilde{\mathrm{Mu}}_{w_i}$ can be written as follows:

\begin{align}  \label{eq:final_tilde_p_new}
&  \tilde{\mathrm{Mu}}_{w_i}\left(z|N,\exp\prod_{j=1}^{J}\left( \mathcal{H}_{w_i}(r_1\wedge ...\wedge r_l|r_{\mathrm{head}}) \right)\right).
\end{align}

The new $D_{\mathrm{KL}}(\tilde{p}_{w_i}(z_i) ||\tilde{p}_{\theta}(\bar{z}))$ also need be re-expressed as:
\begin{align}\label{eq:min_w_add_kl_spec_new} 
\mathrm{Mu}_{\theta}\log\left(\frac{\tilde{\mathrm{Mu}}_{w_i}}{\mathrm{Mu}_{\theta}}\right).\!
\vspace{-2mm}
\end{align}

For that, our new algorithm is described at Algorithm~\ref{ag:algorithm2} as follows:

\begin{algorithm}[!t]
\caption{Fast-\methodname{}}
\label{ag:algorithm2}
\begin{algorithmic}[1]
\STATE Initialize
\FOR{ round $k=0,1,2...,K$ }
\STATE //On each FL Client:
\FOR{ node $i=0,1,2...,n$ }
\STATE Receive shared prior probabilities $T_{\theta}(r_{\mathrm{head}})$ from the FL server to build a rule distribution and sample $J$ unique rule bodies $r_1\wedge ...\wedge r_l$ with it.
\IF {$k < I$}
\STATE Score these candidate rule bodies with Eq.~\eqref{eq:score_f}.
\STATE Update $w_i$ to solve Eq.~\eqref{eq:one_many_FL.b} by minimizing Eq.~\eqref{eq:one_many_FL_lower} by maximizing Eq.~\eqref{eq:log_pw_KG} and using Eq.~\eqref{eq:min_w_add_kl_spec} as a regularization  term.
\STATE Update the new rule's posterior by Eq.~\eqref{eq:final_tilde_p} with ~$w_i^{\star}$.
\ELSE
\STATE Score these candidate rule bodies with Eq.~\eqref{eq:score_f_new}.
\STATE Update $w_i$ to solve Eq.~\eqref{eq:one_many_FL.b} by minimizing Eq.~\eqref{eq:one_many_FL_lower} by maximizing Eq.~\eqref{eq:log_pw_KG_new} and using Eq.~\eqref{eq:min_w_add_kl_spec_new} as a regularization  term.
\STATE Update new rule's posterior by Eq.~\eqref{eq:final_tilde_p_new} with ~$w_i^{\star}$.
\ENDIF
\STATE Send the new rule's posterior to the FL server.
\ENDFOR
\STATE //At the FL server:
\STATE Receive rule posterior probability from clients.
\STATE Generate samples based on posterior probability.
\STATE Use the generated samples to update $\theta$ by maximizing Eq.~\eqref{eq:max_p_theta} to solve Eq.~\eqref{eq:one_many_FL.a}.
\STATE Distribute new shared prior probabilities $T_{\theta}(r_{\mathrm{head}})$ to each client.
\ENDFOR
\end{algorithmic}
\end{algorithm}

In lines 7-9 of Algorithm \ref{ag:algorithm2}, when the current round falls within the initial $I$ steps, we utilize the original graph-based approach $\max_{\mathcal{G}_i}{\prod_{\substack{k=1 \\ r_1\wedge ...\wedge r_l \in z_{ij}}}^{l}x(r_{kj})}$ to calculate the fuzzy value across the entire graph $\mathcal{G}_i$. This uses rule scores to construct Eq.~\eqref{eq:log_pw_KG} and incorporates the KL-divergence constraint (Eq.~\eqref{eq:min_w_add_kl_spec}) as a regularization term to update the weight $w$. The updated $w$ is then used to establish a new posterior distribution by Eq.~\eqref{eq:final_tilde_p}. This approach rectifies the rule bodies using local graph path information, aligning the equivalent paths distributed by the server with the true equivalent paths within $\mathcal{G}_i$, thereby mitigating the need for time-consuming path searches to rectify deviations in specific equivalent paths.

In lines 11-13 of Algorithm \ref{ag:algorithm2}, it is noted that once a foundational and relatively stable rule generator is established, performing a graph search in each round is no longer necessary. Instead, the path distributed by the server is directly utilized, selecting the maximum fuzzy value along the path to effectively represent the score of the current path (${\max_{\substack{k=1 \\ r_1\wedge ...\wedge r_l \in z_{ij}}}^{l}x(r_{kj})}$). This process assists in building a new corresponding Eq.~\eqref{eq:log_pw_KG_new} and incorporates the corresponding KL-divergence constraint (Eq.~\eqref{eq:min_w_add_kl_spec_new}) as a regularization term to update the weight $w$. The updated $w$ is then used to establish a corresponding posterior distribution by Eq.~\eqref{eq:final_tilde_p_new}.

\begin{figure}[h]
\centering
\includegraphics[width=0.92\columnwidth]{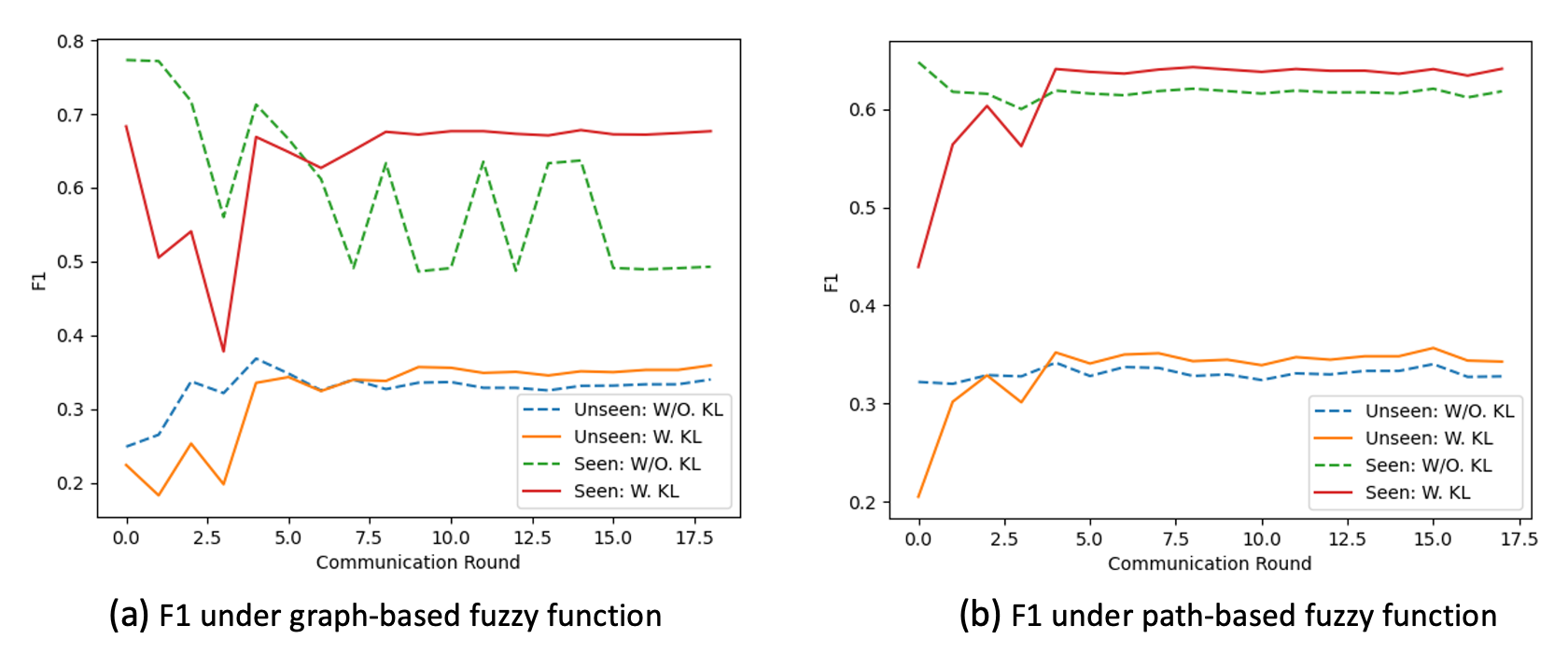}
\caption{Path-based score function has a better effect on reducing the fluctuations than the graph-based score function.}
\label{fig:diff_algo}
\end{figure}

Algorithm \ref{ag:algorithm2} introduces two notable advantages:

\begin{itemize}
    \item \textbf{Reduce fluctuations between upper-lower levels}: The algorithm effectively avoids excessive specializations when the lower level updates the fuzzy values along the trajectory of the rule body of server-distributed rules. Graph-based fuzzy value updating method will result in calculating the fuzzy score totally depending on local specific information which can lead to volatile overall performance fluctuations. As illustrated in Figure~\ref{fig:diff_algo}, plot (b) showcases the F1 values using Algorithm \ref{ag:algorithm2}, achieving nearly identical performance and reducing fluctuations between upper-lower levels in both seen and unseen data scenarios.
    
    \item \textbf{Reduce time complexity}: This algorithm significantly reduces time complexity by only calculating fuzzy values along the trajectory of the rule body but not updating fuzzy values across the graph. More details are in the section on Computational Complexity Analysis for Cross-Domain Learning.
\end{itemize}

\subsection{Computational Complexity Analysis}

In this section, we present an example to compare different computational complexity with different ways to achieve KG-graph information communication. As depicted in Figure~\ref{fig:two_flow} (b), consider the relation $a$ in the triplet $<5,a,6>$ as an instance. Our objective is to enhance the prediction accuracy of relation $a$ by identifying more new rule bodies (equivalent paths from node 5 to node 6) to infer and update the fuzzy value of rule head $a$. In domain 1, there are two equivalent paths from node 5 to node 6, $(b,c)$ and $(e,a,c)$, while in domain 2, paths $(b,c)$ and $(f,h)$ include a distinct path $(f,h)$ that offers new information from another domain. Increasing sample diversity is known to reduce error rates and improve relationship prediction accuracy.

To update all relation edge prediction values across these two domains by discovering new equivalent edges, we consider four approaches:

\begin{itemize}
    \item \textbf{Deterministic Graph Communication}: In this baseline approach, the client-side time complexity does not require path searching. On the server side, adjacency matrices and fuzzy values from both domains are transmitted to the server. The server then merges these matrices to create a larger dimensional adjacency matrix with corresponding relations. The most time-consuming phase is searching for paths within the merged graph space, seeking correspondences between entities from different graphs, represented by $\phi = \{ (e_{1i}, e_{2j}) \mid e_{1i} \in E_1, e_{2j} \in E_2 \}$, resulting in a search space size $S_{KG} = |E_1| \times |E_2|$, which is enormous due to the vast size of entities.

    \item \textbf{Deterministic Rule Communication}: In this approach from \cite{zhang2023lr}, the client-side time complexity involves an intra-domain maximum weight path search for each domain using a uniform NER-pair query pair as endpoints to find all rules. On the server side, the resulting paths are merged based on the uniform NER-pair query pair, aligning path rules between systems ($\psi = \{ (r_{1i}, r_{2j}) \mid r_{1i} \in L_1, r_{2j} \in L_2 \}$), leading to a search space size $S_{LR} = |L_1| \times |L_2|$. While rule alignment provides a better search space scenario compared to entity alignment, it is deterministic, and any graph change necessitates re-alignment.

    \item \textbf{Stochastic Rule Distribution Communication (Algorithm~\ref{ag:algorithm})}: In Algorithm~\ref{ag:algorithm}, the client-side requires an intra-domain path search for scoring candidate rules, confined to paths distributed by the server. The server learns a rule distribution for alignment, a stochastic method ($\theta = \{ (d_{1i}, d_{2j}) \mid d_{1i} \in D_1, d_{2j} \in D_2 \}$), where $|D_1|$ and $|D_2|$ denote the number of distributed rules. The search space size $S_{RD} = |D_1| \times |D_2|$ is significantly smaller than $S_{LR}$ and $S_{KG}$, involving only communication with distribution probability.

    \item \textbf{Stochastic Rule Distribution Communication (Algorithm~\ref{ag:algorithm2})}: For Algorithm~\ref{ag:algorithm2}, the client side only requires a limited intra-domain path search during the initial $I$ rounds. The server's time complexity is the same as in Algorithm~\ref{ag:algorithm} and much less than the two deterministic methods.
\end{itemize}

\subsection{Sensitivity of Adding Ratio of Posterior Sample for Upper-Level Training}

\begin{figure}[h]
\centering
\includegraphics[width=0.72\columnwidth]{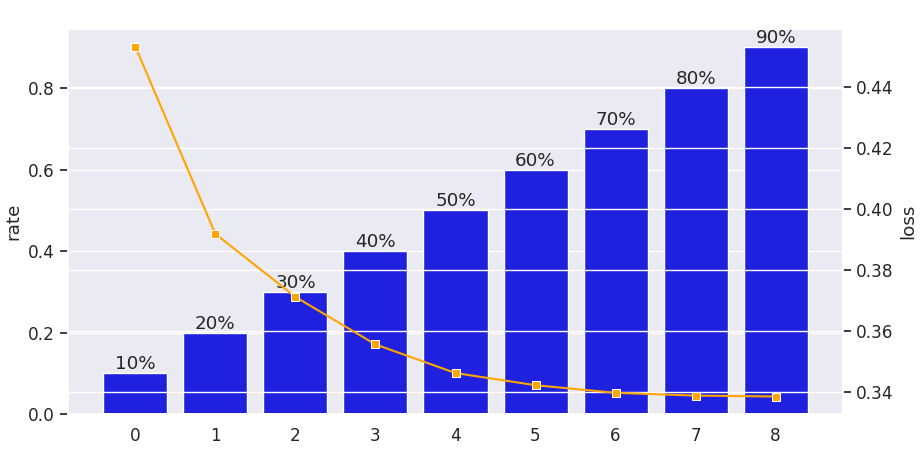}
\caption{Different upper-level first-round training loss when adding different rates of posterior sample.}
\label{fig:join_post}
\end{figure}

To examine whether the incorporation of posterior samples obtained from the lower-level (E-step) can effectively impact the upper-level rule updates in the M-step, we conducted a total of 9 experiments ranging from a sample inclusion ratio of $10\%$ to $90\%$ as shown in the Figure~\ref{fig:join_post}. This comparison aimed to assess the varying effects of different ratios on the training loss of the upper-level rule trainer during its first round of training.

The results indicate that a mere increase in the inclusion ratio of posterior samples from $10\%$ to $20\%$ has a substantial effect in reducing the training loss of the upper-level model. Subsequently, as this ratio is uniformly raised, the loss continues to decrease, albeit at a progressively slower rate. The improvement in the upper-level model becomes less pronounced after incorporating around $70\%$ of the posterior samples. This phenomenon provides evidence for the effectiveness of posterior samples in enhancing the upper-level model. The limited impact observed with higher ratios is consistent with the principle that a certain proportion of samples can effectively reflect the contained posterior information; an excessive number of samples could lead to information redundancy.

\subsection{Sensitivity of Upper-Level Learning Rate}

In this experiment, we conducted tests using three sets of different learning rates for the upper-level rule learner under two conditions: with KL-divergence constraint and without KL-divergence constraint. These tests were performed on both seen and unseen test data to evaluate the impact of the rule generator on the overall game system. As depicted in Figure~\ref{fig:diff_up_lr}, we observe three phenomena:

\begin{figure}[h]
\centering
\includegraphics[width=0.75\columnwidth]{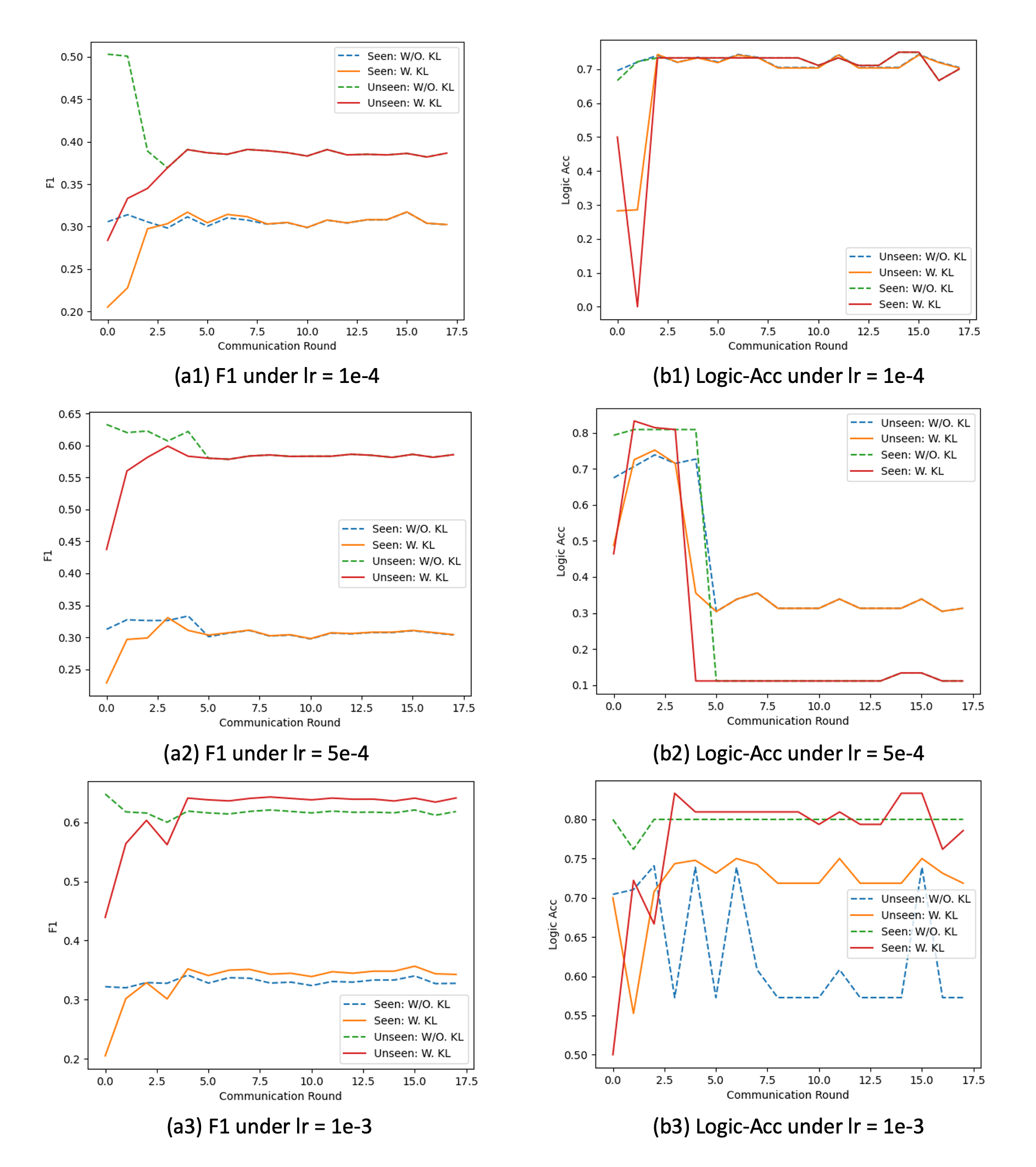}
\caption{F1 and Logic Acc curves under different rule generator’ learning rates.}
\label{fig:diff_up_lr}
\end{figure}

\begin{itemize}

    \item The first phenomenon is that KL dominance occurs only when the upper level achieves dominance in the upper-lower level game, i.e., a larger upper-lower learning rate. This observation aligns with our formula's implication that when the lower level's specialization abilities are in agreement, the group with stronger generalization ability can learn more global prior knowledge at the given learning rate. Consequently, this leads to enhanced overall performance. In cases of insufficient learning rate, the upper-level learning fails to acquire adequate information about the global prior to using the global generalization ability.

    \item The second phenomenon is that while the overall F1 value and the overall logic value increase as the upper learning rate increases, the curves of logic accuracy and F1 only exhibit greater consistency when the learning rate is sufficiently high. This finding indicates that the upper level is indeed learning effective rules to improve the overall F1 performance. When the rule learner fails to learn sufficiently, the rules newly distributed from the upper level may not necessarily be effective for the lower level.

    \item The third phenomenon is that the gap between the curves corresponding to the two groups, unseen and seen, is smaller when the upper layer model employs a lower learning rate. This effect is evident when the upper layer is disadvantaged in the game between the upper and lower layers. This behavior can be observed in both subplots of the F1 values corresponding to (a1) and the logic accuracy values corresponding to (b1), with the latter curve in (b1) almost overlapping. This phenomenon implies that the upper model can differentially generate rules for the unseen and seen data, and as this role weakens, the difference between the curves of seen and unseen data becomes smaller.

\end{itemize}

\end{document}